\pdfoutput=1

\PassOptionsToPackage{table}{xcolor}
\documentclass[11pt]{article}

\usepackage{acl}

\usepackage{tabularx}
\usepackage{subcaption} 
\usepackage{placeins}
\usepackage{makecell}
\usepackage{xcolor}
\usepackage{xspace}

\newcommand{\name}{\textsc{DeM-MoE}\xspace}

\usepackage{amsfonts} 

\newcommand{\myparagraph}[1]{\noindent \textbf{#1}}

\newcommand{\camerareadytext}[1]{\xspace}

\newcommand{\fref}[1]{Figure~\ref{#1}}
\newcommand{\tref}[1]{Table~\ref{#1}}

\usepackage{times}
\usepackage{latexsym}
\usepackage{breqn}
\usepackage{booktabs} 

\usepackage[T1]{fontenc}

\usepackage[utf8]{inputenc}
\usepackage{amsmath} 

\usepackage{microtype}
\usepackage{booktabs}
\usepackage{adjustbox}
\usepackage{longtable}

\usepackage{inconsolata}
\usepackage[most]{tcolorbox}

\usepackage{graphicx}
\usepackage{etoolbox}
\makeatletter
\pretocmd{\thebibliography}{%
  \setlength{\parskip}{0pt}
  \setlength{\itemsep}{0pt plus 0.3ex}
  \setlength{\parsep}{0pt}%
  \setlength{\partopsep}{0pt}%
}{}{}%
\makeatother

\title{Modeling Annotator Disagreement with Demographic-Aware Experts and Synthetic Perspectives}

\author{
Yinuo Xu${}^{1}$, Veronica Derricks${}^{2}$, Allison Earl${}^{1}$, and David Jurgens${}^{1}$ \\
${}^{1}$University of Michigan \\
${}^{2}$University of Colorado Boulder \\
\texttt{\{yinuoxu, anearl, jurgens\}@umich.edu}, \texttt{Veronica.Derricks@colorado.edu}
}

\DeclareGraphicsExtensions{.pdf,.png,.jpg}
\begin{document}
\maketitle
\begin{abstract}
We present an approach to modeling annotator disagreement in subjective NLP tasks through both architectural and data-centric innovations. Our model, \name (Demographic-Aware Mixture of Experts), routes inputs to expert subnetworks based on annotator demographics, enabling it to better represent structured, group-level variation compared to prior models. \name consistently performs competitively across demographic groups, and shows especially strong results on datasets with high annotator disagreement. To address sparse demographic coverage, we test whether LLM-generated synthetic annotations via zero-shot persona prompting can be used for data imputation. We show these synthetic judgments align moderately well with human annotations on our data and offer a scalable way to potentially enrich training data. We then propose and evaluate approaches for blending real and synthetic data using strategies tailored to dataset structure. We find that the optimal strategies depend on dataset structure.  Together, these contributions improve the representation of diverse perspectives\camerareadytext{ in subjective NLP}.
\end{abstract}

\section{Introduction}

Substantial disagreement among annotators is common in subjective tasks such as toxicity detection, misinformation labeling, and politeness evaluation. These disagreements often reflect meaningful differences in perspective rooted in social, cultural, or demographic backgrounds rather than random noise \cite{holland1987cultural, larimore-etal-2021-reconsidering,sap-etal-2022-annotators}. For example, a comment judged toxic by younger users might seem benign to older ones, reflecting differing sensitivities to language and topics. Modeling such perspective variation is essential for building systems that represent and reason over diverse viewpoints. Yet many approaches treat disagreement as noise, collapsing annotations into a single label and marginalizing minoritized perspectives \cite{prabhakaran-etal-2021-releasing}. Recent work instead treats disagreement as signal \cite{dawid1979maximum}, using weighting, filtering, or learning from annotation distributions \cite{10.1613/jair.1.12752}. Distributional modeling has become prominent in NLP tasks where disagreements reflect socially grounded variation \cite{davani-etal-2022-dealing,fleisig2024majority,Gordon_2022,Wan2023}. However, such methods often lack inductive biases to capture structured group-level variation, risking underrepresentation of marginalized perspectives.

We introduce \name, a Demographic-Aware Mixture of Experts model that learns to represent subjective judgments by routing inputs to expert subnetworks based on annotator demographics. This design introduces inductive bias: similar annotators may reason about inputs similarly, encouraging specialization and improving representation of underrepresented groups. \name outperforms strong baselines and SOTA systems across multiple datasets and demographic groups. Beyond architectural innovation, we address modeling disagreement in low-data settings with sparse demographic coverage using LLM-generated synthetic annotations and blended training strategies. Optimal strategies depend on dataset structure, showing how \name, combined with strategic data augmentation, effectively models viewpoint diversity even in low-resource scenarios.

We make three contributions: (1) We propose \name, a modular architecture with demographic-aware routing to capture structured variation in annotation behavior. (2) We evaluate the alignment of zero-shot LLM-generated annotations with human ratings. (3) We present a framework for incorporating synthetic data, showing that its effectiveness depends on dataset structure.\footnote{Data and code are available at \emph{http://anonymized}.}

\section{Related Work}

\myparagraph{Modeling Individual \& Group Preferences in Recommendation Systems.}
Recommendation systems extensively model individual and group preferences. Group-based methods aggregate member choices into collective decisions, often using centrality-aware representations \cite{9195784} or hierarchical attention to separate individual and group signals \cite{Xu2024AlignGroupLAA,Liang2022AHAA,Wang2024BiLevelUMA}. Mixture-of-Experts (MoE) architectures further capture multifaceted user interests \cite{Liu2024FacetAwareMMA}, model user–item interactions \cite{Zhao2020DoubleWingMOA}, and enable multi-task personalization \cite{Kong2024CustomizingLMA}, including user and group modeling \cite{gong2023attentionweightedmixtureexperts,liu2025mixtureofexpertspersonalizedsemanticawarelocation}. In contrast, no prior NLP work uses MoEs to jointly model individual and group variation. While our task and recommenders model variation across individuals and groups, the nature and objectives differ. Whereas recommenders target idiosyncratic preferences with weak demographic patterns and aim to merge signals into a unified ranking across items, our task focuses on predicting singular judgments using systematic socio-demographic regularities. Annotators from similar groups may evaluate content in shared ways, and rather than collapsing disagreement, our approach explicitly preserves them to represent diverse perspectives.


\myparagraph{Modeling Annotator Disagreement in NLP.} Disagreement among annotators is common in subjective NLP tasks like toxicity classification, misinformation detection, and stance analysis. Traditional methods treat disagreements as noise by using majority voting or averaging to create a single "gold" label per instance, which can obscure meaningful variation from underrepresented or minoritized groups \cite{prabhakaran-etal-2021-releasing}. Alternatives include early work that measured or filtered disagreement to improve data quality \cite{aroyo2015truth, reidsma-op-den-akker-2008-exploiting, beigman-klebanov-beigman-2014-difficult}, and more recent approaches that learn from disagreement directly:
a) Uncertainty-based methods, which weight examples by annotation variability \cite{plank-etal-2014-learning};
b) Distributional and multi-task models, which use label distributions or treat annotators as tasks \cite{davani-etal-2022-dealing};
c) Annotator modeling frameworks, which predict individual labels using shared encoders and per-annotator heads or embeddings \cite{fleisig2024majority, Gordon_2022, Wan2023}. Recent work emphasizes the role of demographics in annotation behavior. \citet{Wan2023} model demographics to predict disagreement, while \citet{Gordon_2022} use a jury learning approach to estimate group verdicts. However, most models treat demographic features as input to shared encoders, without architectural diversity or inductive biases to model group-specific reasoning. As a result, they capture individual variation but struggle with \camerareadytext{systematic} group-level differences.

\begin{table}[t]
\centering
\rowcolors{2}{gray!8}{white}

\resizebox{0.48\textwidth}{!}{%
\small
\begin{tabular}{@{}lrrrrrl@{}}
\toprule
\textbf{Dataset} & \textbf{\#Inst} & \textbf{\#Ann} & \textbf{Avg/Inst} & 
\textbf{IAA ($\alpha$)} & \textbf{Entropy} & \textbf{Max Coef} \\
\midrule
Safety      & 350 & 123 & 123.0 & 0.241 & 0.742 & Race (0.559) \\
Offensiveness  & 1,500 & 262 & 8.69 & 0.287 & 1.212 & Age (1.351) \\
Patient Centered Comm. & 2,230 & 589 & 3.33 & 0.287 & 1.492 & Age\_group (2.528) \\
Politeness    & 3,718 & 506 & 6.74 & 0.440 & 1.395 & Race (1.427) \\
Toxicity     & 107,620 & 17,172 & 4.74 & 0.272 & 1.070 & Age\_range (2.056) \\
\bottomrule
\end{tabular}%
}
\caption{Dataset statistics. `\#Inst'' = number of instances, \#Ann'' = annotators, Avg/Inst'' = avg.\ annotators per instance, IAA'' = Krippendorff’s $\alpha$, Entropy'' = mean entropy of annotator ratings per instance. ``Max Coef'' = demographic feature with the largest absolute ridge regression coefficient per dataset. }
\label{tab:condensed_dataset_stats}
\end{table}

\section{Data}
We evaluate on five datasets spanning diverse tasks, demographic coverage, and levels of disagreement (Table~\ref{tab:condensed_dataset_stats}; see full details in Appendix~\ref{sec:data-overview}). The Toxicity dataset \citep{kumar2021designing} is the largest, with the broadest demographic coverage (2,535 combinations) and low agreement ($\alpha$=0.27), suggesting systematic disagreement. Ridge regression predicting annotator ratings using demographics shows strong demographic signal, especially for age (2.06). The Safety dataset \citep{aroyo2023dicesdatasetdiversityconversational} contains dense annotations (123 per instance) across 48 combinations, but has low agreement ($\alpha$=0.24). Demographic signal is weak likely due to the complexity of harmfulness judgments. The \textsc{Popquorn} dataset \cite{pei2023annotatordemographicsmattermeasuring} includes Politeness and Offensiveness. Politeness shows the highest agreement ($\alpha$=0.44) and low entropy, with strong demographic effects (race\camerareadytext{and education}). Offensiveness\camerareadytext{with moderate demographic diversity and agreement} shows notable signal for age and occupation. The Patient Centered Communication (PCC) dataset rates doctor–patient interactions across multiple attributes\footnote{Details on how PCC is collected are in Appendix \ref{sec:pcc-details}}. It has sparse annotations (3.3 per instance) and high uncertainty (entropy). PCC also shows the strongest demographic signal, especially for age. These datasets highlight varying balances of idiosyncratic vs. systematic disagreement, underscoring the need for models that preserve demographic variation.

\section{\name for Modeling Disagreement}

Recent work on annotator disagreement has moved beyond majority-vote aggregation toward models that predict annotator-specific labels \cite{davani-etal-2022-dealing,fleisig2024majority,Gordon_2022}. Many incorporate annotator identity or demographics, but typically by concatenating demographic features with text \cite{fleisig2024majority,Gordon_2022,Wan2023} or by implicitly encoding annotator perspectives through embeddings \cite{deng-etal-2023-annotate}. Because these systems model all annotators through a single network, they cannot explicitly specialize in distinct judgment patterns across demographic groups. Consequently, their ability to represent shared group behaviors remains limited. \citet{orlikowski-etal-2023-ecological} partially addresses this by injecting demographic information into a shared encoder with group-specific layers, but intersectional identities are still modeled through separate, independent components, constraining the model’s capacity to learn intersectional patterns.

We propose a new approach, \name (Demographic-aware mixture of experts), based on Mixture of Experts (MoE) (Fig. \ref{fig:exp1_moe}) \cite{shazeer2017outrageouslylargeneuralnetworks}, which naturally supports modular specialization (different experts learn distinct annotation patterns linked to demographic groups) and selective routing (inputs are dynamically directed to relevant experts based on annotator demographics). The input consists of the text snippet (encoded with Modern-BERT), the annotator embedding, and the demographic embeddings, and then all three are concatenated into the MoE input, which is then routed to experts to predict the annotator’s rating. Our architecture encodes inductive bias: annotators from similar demographic groups may share systematic ways of judging texts. We address a key gap in prior work: the lack of structured inductive bias. While large networks can learn subgroup variation, they may not do so in a structured or interpretable way. They may also overfit to dominant groups without an architectural signal promoting subgroup differentiation. Our model makes this inductive bias more robust and interpretable, especially under data imbalance or sparse subgroup representation. Our model's routing also allows experts to naturally specialize in intersectional subgroups, and learn cross-group and within-group specialization. We discuss the model components next: 1) learned annotator and demographic embeddings; 2) expert selection and dynamic routing; and 3) expert load balance and specialization via a weighted training loss.

\myparagraph{Annotator \& Demographic Embeddings.} 
We initialize Bayesian embeddings \cite{vilnis2015word} for annotators and their demographic attributes, enabling the model to capture annotator-specific idiosyncrasies and demographic-related biases. Each is represented by a learned Gaussian posterior distribution, with embeddings sampled during training via the reparameterization trick \cite{kingma2014auto}. We concatenate the text (with Modern-BERT \cite{modernbert}), annotator, and demographic embeddings into the MoE input:
\(
\mathbf{x} =
[\mathbf{e}_{\text{text}};\mathbf{e}_{\text{ann}};\mathbf{e}_{\text{demo}}]
\)

\myparagraph{Expert Selection.}  
To promote expert specialization aligned with demographic-group preferences, we build a pool of \(n\) experts ($n=$ number of demographic groups). A lightweight gate maps the input vector to expert scores:
\(
  \mathbf{s} = W_s\,\mathbf{x} + \mathbf{b},
  \mathbf{p} = \mathrm{softmax}(\mathbf{s})
\). We hard-select the top $k$ experts (\(k\in \{2,3\}\)):
\(
  \mathcal{I}_k = \operatorname{arg\,topk}(\mathbf{p},k)
\).
Gating weights \(p_j\) determine which demographic-aware experts are used. Each expert \(f_j\) receives the same input; their outputs are mixed sparsely:
\(\mathbf{h}\;=\;
  \sum_{j\in\mathcal{I}_k} p_j\,f_j(\mathbf{x})
\). A linear regression head predicts the snippet rating. Ratings are z-score normalized to stabilize training. We prefer hard expert selection over soft gating to reinforce specialization\camerareadytext{,  whereas soft gating dilutes gradients}.

\myparagraph{Expert Load Balance \& Specialization.} Naive routing often leads to expert load imbalance, routing collapse, or bottlenecks. While some auxiliary losses improve hardware efficiency via uniformity \cite{Fedus2021SwitchTS, wang2024auxiliarylossfreeloadbalancingstrategy}, we focus on expert diversity with a normalized standard deviation loss, allowing roughly even usage, rather than strict uniformity, to support demographic specialization. We also include an orthogonality loss to encourage distinct expert features, and a variance loss to promote diverse routing paths \cite{guo2025advancingexpertspecializationbetter}. Our training loss further encodes the inductive bias, where we include regularizations for annotator and demographic embeddings, load loss, orthogonality, variance loss, and demographic within-group specialization loss (see Appendix \ref{app:train_loss}).

\section{Experiment 1: Modeling Perspectives}

 We hypothesize that \textbf{H1}: Models that incorporate demographic structure as an inductive bias more effectively capture diverse annotator perspectives. We first test if the experts are sufficiently specialized within and across demographic groups. We then compare \name to other models' representativeness across demographic groups.

\subsection{Experiment Setup}
All models are evaluated to predict individual annotator ratings using the text of each item and, where applicable, annotator or demographic information. Each model differs in its inputs (text only vs.\ text + annotator/demographics), while the output is a scalar rating per annotator–item pair, except for ModernBERT (which predicts at the snippet level with text input). (1) Probabilistic Matrix Factorization (PMF) \citep{10.5555/2981562.2981720} uses only annotator and item IDs, learning latent embeddings for both and modeling their interaction as a probability distribution over ratings, thus capturing annotator regularity without text or demographics. (2) As a text-based baseline, we fine-tune ModernBERT-large \citep{modernbert} with a single linear regression head on the CLS embedding to predict the mean snippet rating.\camerareadytext{ModernBERT \citep{modernbert} is a text-based baseline that encodes the snippet, and we regress on the embedding to predict the mean snippet rating.} (3) LLaMA-3.1-8B-Instruct \citep{orlikowski2025demographicsfinetuninglargelanguage} receives text alone in a zero-shot setting or text plus annotator demographic descriptions when LoRA-fine-tuned with sociodemographic prompts. Their results improved over text-only models but revealed the model did not benefit from demographics. (4) Annotation + Annotator Embedding model (En + Ea) \citep{deng-etal-2023-annotate} is a SOTA system for
explicitly modeling both item and annotator-level
variance but does not incorporate demographic information. (5) Jury Learning \citep{Gordon_2022} uses ModernBERT text embeddings, annotator embeddings, and group-level demographic embeddings, concatenated and passed through cross and deep networks \citep{Wang_2021} to predict annotator ratings; this architecture most closely parallels ours.

\noindent \textbf{Training.} Train/dev/test splits are created at the instance level, ensuring no overlap of snippets between the splits while allowing some overlap in annotators between the train and test sets. Offensiveness has 92\% overlap, Politeness 91\%, Safety 85\%, Toxicity 40\%, and PCC 87\%. This evaluation focuses on how well the model generalizes to new content and unfamiliar annotators. Training details are in Appendix \ref{exp1op}.

\begin{figure*}[!t]
\centering
\includegraphics[width=0.7\textwidth]{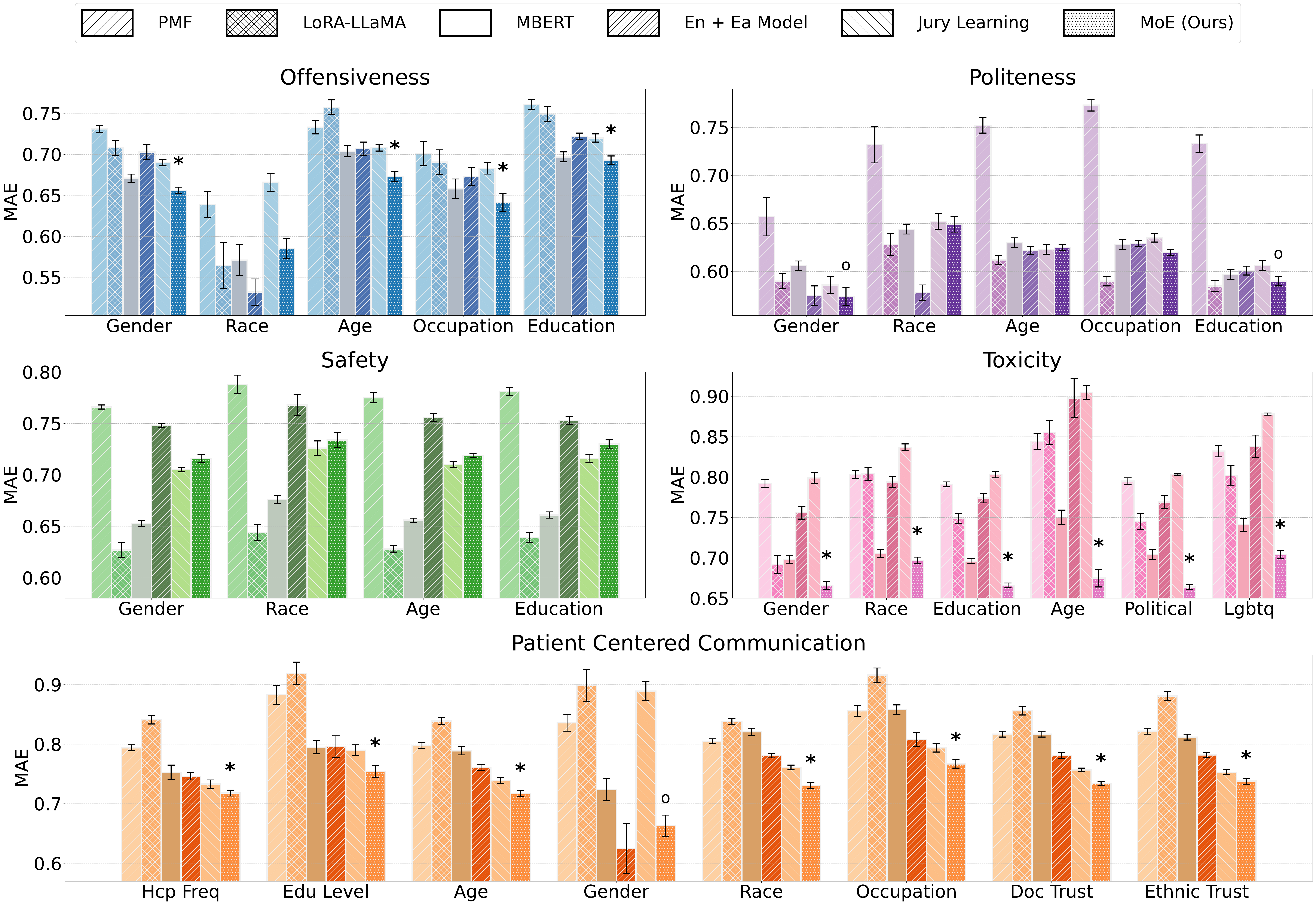}
\caption{Comparison of Mean MAE across demographics for all datasets (lower MAE is better). We obtain the mean and error bars from bootstrap samples. The star (*) above our MoE model indicates that it is statistically better ($p < 0.05$) than next-best model. The circle (o) above MoE indicates that it is statistically equivalent to best model.}
\label{fig:agg_exp1_demo}
\end{figure*}

\noindent \textbf{Evaluation.}
We use Mean Absolute Error (MAE) to measure the average absolute different between the model's predicted rating and actual annotator rating, at the individual level\cite{Gordon_2022, fleisig2024majority}. For analysis, we aggregate across annotators' demographic groups.


\subsection{Results}
\name consistently performs competitively across demographic groups, with particularly strong results on datasets with high annotator disagreement: Toxicity, Offensiveness, and PCC (Figure~\ref{fig:agg_exp1_demo}).\footnote{We also test best zero-shot performance with LLaMA 3.1-8B-Instruct, but omit it from the figure due to its low performance. Results are reported in Appendix \ref{app:best-zero}} Our model outperforms all other models in every demographic group on Toxicity, the most demographically diverse data with low annotator agreement. Our model outperforms all other models in every group except for race on Offensiveness, and gender (but statistically equivalent to the best model) on PCC. On Politeness, it is statistically tied with the best models for gender and education.

Two other trends merit noting. First, no other system consistently performs well across all datasets. For example, PMF (which uses no text or annotator information) generally performs worst. However, it still outperforms LLaMA on PCC, possibly because PMF captures stable annotator-specific preferences statistically. In contrast, the LLM may rely on coarse demographic priors or stereotypes, which are less effective in subjective domains. Similarly, while Jury Learning is often among the second-best approaches, it performs much worse on the Toxicity dataset. Each of these datasets contains unique sources of label variation due to the interactions between content, identity, and demographics; for example, variation in some datasets may be driven more by individual annotators’ preferences rather than by group-level demographic effects. This variation in performance underscores the need to test models across multiple datasets in order to assess their sensitivity to different sources of variation.

Second, among models that use both annotator identity and demographics, we see a trend that increasing model structure generally benefits performance and supports our hypothesis \textbf{(H1)} that models with stronger inductive biases can better learn regularity in label variation. LLaMA has the least structure, encoding identity and demographics as text and learning preferences using next-token prediction. Similar to \citet{orlikowski2025demographicsfinetuninglargelanguage}, we find that their model struggles to capture demographic variation on datasets with strong demographic signals (Offensiveness, Politeness, PCC). However, LLaMA performs best on the Safety task, where demographic influence is minimal. This suggests that it excels in text-dominant settings with limited demographic variability (and benefits from having roughly 1M more parameters than our model). In contrast, \name is worse and likely underfits in such settings, as the routing function provides limited benefit when individual and group preferences are weak or absent. However, more structured models like \name offer greater robustness across a broader range of labeling conditions. The consistently high performance of \name in settings with high demographic signal suggests that expert routing provides a more effective inductive bias than the dense cross-network architecture and a strong capacity to represent fine-grained differences in annotator viewpoints.

While larger model capacity might explain performance boost, we observe this is not the case (Tab. \ref{tab:param_counts}). Models with far larger parameter counts i.e. LoRA-LLaMA (3.4M) often underperform MoE (2.5M), suggesting that our advantage stems from inductive bias rather than increased capacity.

\begin{figure}[tb!]
    \centering    \includegraphics[width=\columnwidth]{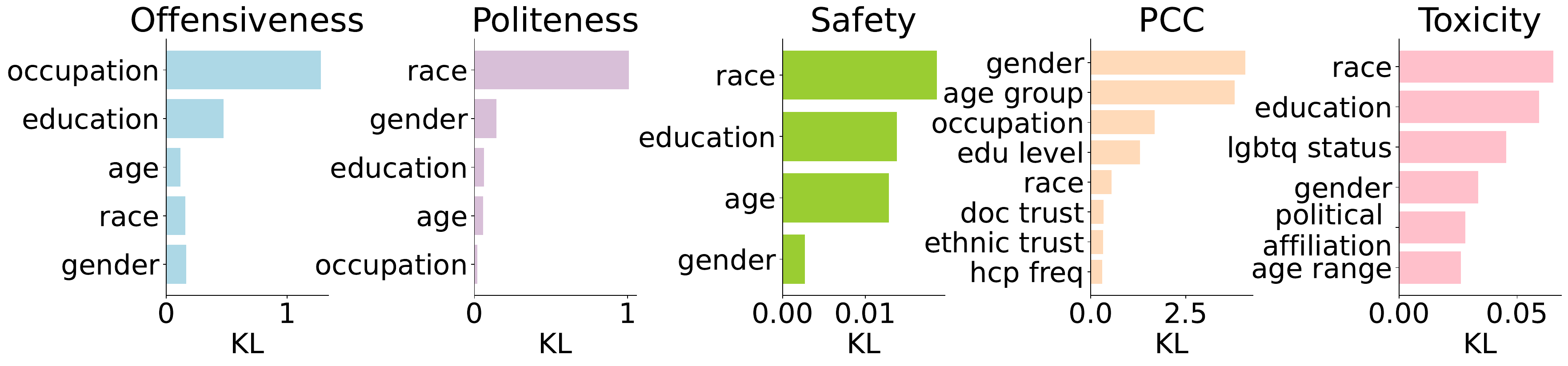}
    \caption{Mean pairwise KL diversity in expert usage distributions across subgroups for each demographic (higher KL shows more specialization).}
    \label{fig:within-group}
\end{figure}

\noindent\textbf{Do Experts Align with Demographics?} To test whether the model’s inductive biases lead to experts aligning with demographic groups, we analyze two types of specialization: within-group and cross-group. Within-group specialization focuses on diversity among subgroups within a demographic (e.g., different racial identities), reflecting our view that not all perspectives within a group can or should be collapsed into one. We quantify this by computing the mean pairwise KL divergence in expert usage distributions across subgroups for each demographic. To normalize for model capacity, we divide each KL score by $\log(K)$, $K$ = the number of experts. We find that experts specialize in capturing variation within demographic groups (\fref{fig:within-group}), but the relevant groups vary by dataset --highlighting interactions between the construct being modeled and demographic-specific variation. The model shows the most specialization on Politeness for race, the strongest predictor of rating variance \cite{pei2023annotatordemographicsmattermeasuring}. Heatmaps of expert usage, aggregated across all instances within each demographic category, show different experts specializing in subgroup perspectives (Figures~\ref{fig:pol-within-group}–\ref{fig:pcc-within-group}).\footnote{The diffuse activation patterns arise because averaging across instances makes multiple experts appear active, reflecting subgroup variability rather than soft gating.} MoE routing provides more granular modeling of demographics meaningfully tied to prediction.

Cross-group specialization asks whether certain experts attend more strongly to specific demographic perspectives overall. We use ridge regression to predict expert usage from demographic attributes and visualize the coefficients in a clustered heatmap (Fig.~\ref{fig:pol-cross-group}–\ref{fig:safety-cross-group}). We observe distinct expert specializations, especially for datasets with high demographic signal (Politeness, Offensiveness, PCC). These analyses show that the MoE learns both fine-grained subgroup distinctions and broader demographic-aligned expert roles, reflecting encoded inductive bias. MoE specialization also reveals the type of demographic variance present in the data. This inductive structure primes MoE to better capture diverse annotation patterns.


\noindent \textbf{What kind of data influences performance?} 
The effectiveness of \name is closely tied to the dataset's properties. Three key factors mediate  performance: (1) Low agreement: MoE performs best in tasks with low IAA (Offensiveness, Toxicity, PCC), where subjective interpretations vary\camerareadytext{ widely}. In such settings, expert specialization provides a clear advantage over models that collapse annotations into a single label. In contrast, tasks with high IAA (Politeness) offer less room for modeling perspective diversity, limiting the relative benefits of MoE. \camerareadytext{ MoE excels when there is item-level variance, i.e., per-instance entropy and standard deviation are high (PCC, Offensiveness, Politeness), allowing experts to model ambiguity in a structured way. Even in datasets like Safety, which have low entropy but extremely dense annotation per instance, MoE still benefits by capturing diverse views at the instance level despite lower overall ambiguity.}(2) Demographic predictiveness of ratings: MoE performs best when demographic attributes are predictive of annotator ratings (e.g., age and gender in PCC\camerareadytext{, occupation and age in Offensiveness}). Expert specialization aligns well with this variance, as performance is weaker when demographics are less predictive (Safety). (3) Annotation density: MoE benefits from having sufficient annotations per demographic profile to support expert learning. Demographic error was correlated ($r$=0.40) with the number of annotations for that group; where datasets like Toxicity with many annotations per demographic combination (213) can be modeled more effectively than those like Politeness with lower density (44).


\camerareadytext{In summary, \name is most effective when there is: high annotator disagreement (globally or locally), strong demographic signal in ratings, and sufficient annotation density per demographic group.}


\section{Experiment 2: Synthetic Annotations}


Experiment 1 showed that \name can represent labels effectively in settings with dense annotations and diverse annotators. Yet obtaining such data is costly and logistically difficult. Prior work suggests LLMs can approximate group-specific perspectives without training \cite{beck-etal-2024-sensitivity, hu-collier-2024-quantifying}. We hypothesize that \textbf{(H2)} LLM-generated annotations conditioned on demographic personas can moderately align with human ratings on subjective tasks. Before training \name with synthetic data, we first test LLMs’ ability to capture demographic regularities in our datasets, laying the groundwork for Experiment 3 on combining synthetic and real annotations.

\subsection{Experiment Setup}

We evaluate the zero-shot performance\footnote{Pilot experiments showed that zero-shot generally performed better than few-shot with our data, we use the former. See Appendix \ref{app:few-shot} for details.} of four instruction-tuned LLMs with reasoning capabilities:\camerareadytext{some of which are evaluated in similar tasks in prior work \cite{orlikowski2025demographicsfinetuninglargelanguage, sun2025sociodemographicpromptingeffectiveapproach}}
LLaMA-3.3-70B-Instruct, QwQ-32B, OLMo-2-13B, and Mistral-Nemo-Instruct-2407. As baselines, we include a model that predicts ratings at random, and one that predicts the dataset's mean rating. \camerareadytext{These baselines are simple approaches with no nuanced content analysis or demographic insight.} Each LLM is told to adopt the perspective of a given demographic persona,  provide a short reasoning for the rating, and output the final rating. Full prompt templates are in the Appendix \ref{app:exp20}. We evaluate \camerareadytext{outputs}using Pearson's $r$ for alignment with human labels, and MAE for accuracy.

\subsection{Results}

LLaMA outperforms other models in alignment with human judgments (\tref{tab:exp2_model}). Though the level of alignment varies by task, these results suggest that the model is able to generate predictions that are reasonably aligned in magnitude; an analysis with Pearson's $r$ (Appendix Table \ref{tab:exp2_model_r}) shows the same model trends and confirms that LLM predictions are directionally aligned as well, and reasonably calibrated in magnitude. 
Overall, LLMs struggle with simulating ratings for conversations (Safety and PCC) tasks that have low human annotator agreement in the first place. MAE is lowest for Safety, but this reflects its narrower 1-3 rating scale rather than higher accuracy; all other datasets uses a 1-5 scale. To test the construct validity of the synthetic data for rare demographic groups, we conduct a more granular analysis of the performance of different demographic groups relative to their frequency in the data (App. \ref{app:exp2-group-results}). In general, there isn't a performance gap between the dominant and minoritized groups (though there are a few exceptions, such as PCC annotators with "Other" race, or Politeness annotators with "less than a high school diploma").
These findings support our hypothesis \textbf{(H2)}: zero-shot demographic prompting helps LLMs to approximate human ratings, and could be useful for data imputation to better learn demographic perspectives. Next, we test whether combining synthetic and real annotations improves performance in data-limited settings.


\begin{table}[t]
\small
\setlength{\tabcolsep}{4pt}
\centering
\begin{tabular}{@{}lccccc@{}}
\textbf{Model} & \textbf{OFF} & \textbf{POL} & \textbf{Safety} & \textbf{PCC} & \textbf{TOX} \\
\hline
Random & 0.954 & 1.127 & 0.851 & 1.233 & 1.311 \\
\rowcolor{gray!8}
Mean Predictor & 0.815 & 0.846 & 0.829 & \textbf{ 0.873 } & 1.045 \\
LLaMA-3.3-70B & \textbf{0.778} & \textbf{0.933} & \textbf{0.488} & 1.015 & \textbf{0.927} \\
\rowcolor{gray!8}
OLMo-2-13B & 1.096 & 1.121 & 0.878 & 0.989 & 1.269 \\
Mistral-Nemo & 1.449 & 0.956 & 1.113 & 1.028 & 1.373 \\
\rowcolor{gray!8}
QwQ-32B & 1.068 & 1.350 & 0.553 & 1.071 & 1.178 \\
\end{tabular}
\caption{Mean Absolute Error (MAE) across models and datasets (lower is better). Best scores are bolded.}
\label{tab:exp2_model}
\end{table}



\section{Experiment 3: Model Training with Real and Synthetic Annotations}

The results of Experiment 2 suggest that LLMs can be used to impute moderately-aligned ratings for training.  These synthetic data offer  several potential benefits\camerareadytext{ for our task}:1) new annotations from underrepresented demographic perspectives, helping reduce bias and improve diversity in training \cite{zhezherau2024hybridtrainingapproachesllms, chen2024diversitysyntheticdataimpact, li2024selfalignmentinstructionbacktranslation}; and 2) scalable dataset sizes without the cost and time required for additional human labeling \cite{chan2024balancingcosteffectivenesssynthetic,chung2022scalinginstructionfinetunedlanguagemodels}. However, the benefits of synthetic data depend on careful integration. Poorly aligned or noisy synthetic data can introduce bias, harm generalization, and risk misrepresenting minority group perspectives \cite{wyllie2024fairnessfeedbackloopstraining, shumailov2024curserecursiontraininggenerated, pereira2021analysisdeploymentmodelstrained, pmlr-v162-ganev22a}. The optimal method for combining real and synthetic data remains unresolved. Some work prevents model collapse by training on both original and synthetic data \cite{gerstgrasser2024modelcollapseinevitablebreaking}, while others experiment with pretraining-finetuning schemes or balanced data blending \cite{maini2024rephrasingwebrecipecompute, zhezherau2024hybridtrainingapproachesllms, Krishna2021DoesPF, Doshi2024PretrainingLMA}. However, these works focus using synthetic data to improve task performance, whereas our work uses the data to improve our ability to model the people labeling for the task. 


We present a systematic framework to test configurations of synthetic data generation methods with training strategies to blend real and synthetic data optimally. Our goal is to enhance the performance of \name by increasing the diversity of perspectives. We hypothesize \textbf{H3}: strategic integration of synthetic data into training could improve the task of modeling disagreement, and better representation across various demographics.

\subsection{Experiment Setup}

Experiment 3 tests two aspects of using synthetic data: 1)which synthetic data is generated; 2) how the synthetic data is incorporated during training.


\subsubsection{Generating Synthetic Annotations}
To evaluate the effects of scale and representativeness of synthetic annotations on performance, we compare three quantity-based and one quality-based strategy. We first extract all demographic combinations from the real dataset to build a pool of synthetic personas. 
\textbf{1) Random Strategies:}
For an instance with $n$ real annotations, 
\textbf{0.5x (random):} add 0.5$n$ synthetic annotations;
\textbf{1x (random):} add $n$ synthetic annotations; and
\textbf{Fill (random):} add up to the maximum number of annotations per instance to ensure uniform coverage. All synthetic annotations are generated using randomly-sampled personas. These strategies prioritize increasing the \textit{quantity} of data.
\textbf{2) Non-Random Strategy:} \textbf{Cluster:} Use k-means clustering on real demographic profiles and rating behavior (mean and SD) to select 20 representative annotators near each cluster centroid. To induce disagreement, sample 20 from the most distant clusters. For each instance, a cluster is chosen at random, with half of the synthetic annotators drawn from representatives and half from disagreeers. This process aims to improve \textit{quality} by adding view diversity.

\begin{figure*}[!t]
  \centering
  \includegraphics[width=0.65\textwidth]{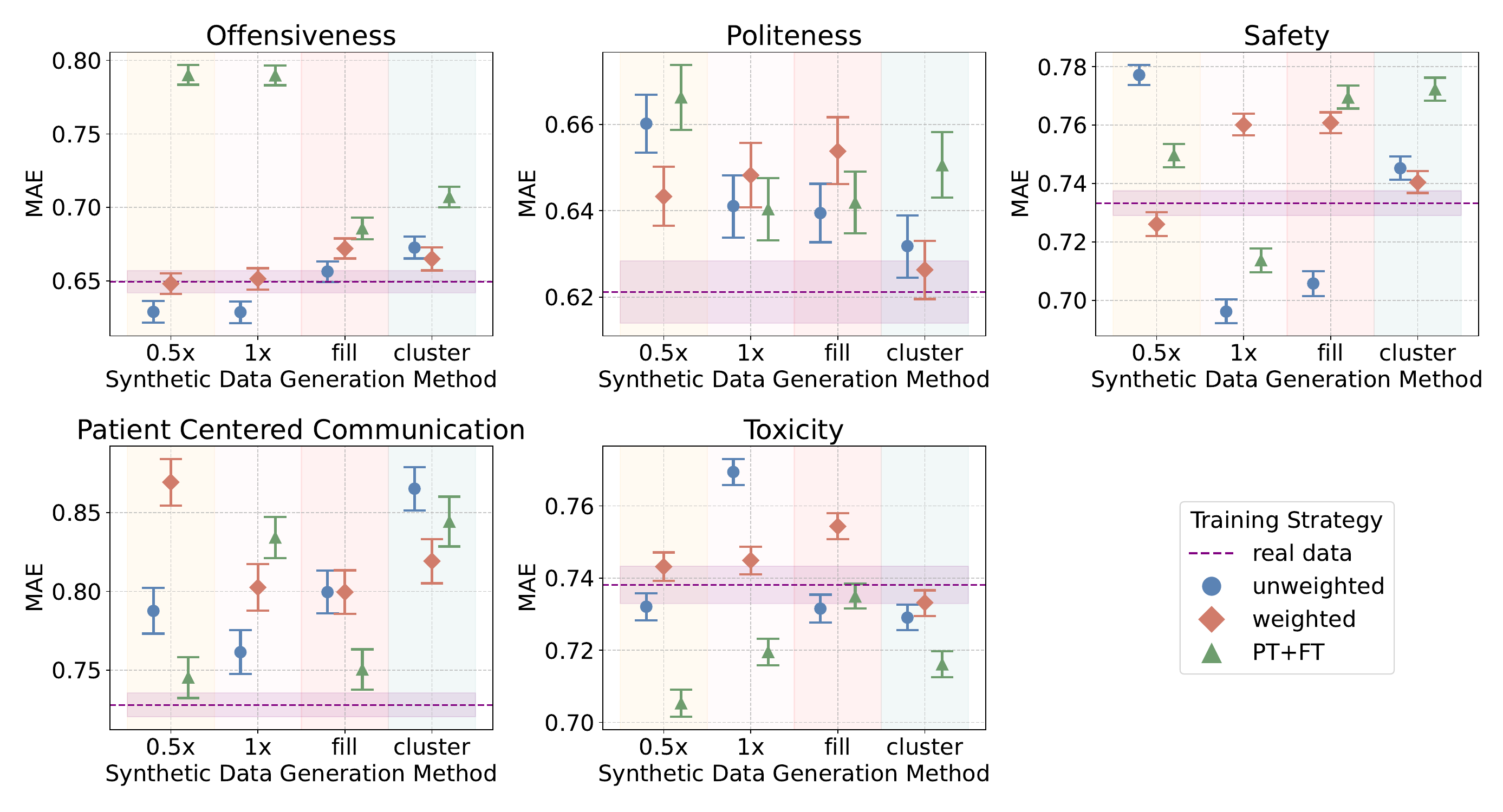}
  \caption{Mean MAE across demographic categories by training strategy and synthetic-data generation method (lower is better), shown for the three datasets. The purple horizontal line is the MAE of \name trained only on real data (see Experiment 1) with 95\% confidence intervals. The shaded regions denote \camerareadytext{different synthetic} data generation methods\camerareadytext{: the warm-hued regions denote random generation, and the teal regions denote non-random cluster generation}.}
  \label{fig:exp3_all}
\end{figure*}

\subsubsection{Blending Real and Synthetic Data\camerareadytext{During Training}}

We consider three strategies for how to incorporate synthetic annotations during training. (1) \textbf{Pretrain and Fine-Tune} (PT+FT) pretrains the model using only synthetic data and the fine-tunes with real data. This approach aims to learn general patterns from synthetic data and refine them using real annotations. (2) The \textbf{Unweighted} strategy mixes the real and synthetic data during training time, treating mistakes on either dataset equally. (3) The third strategy recognizes that unweighted training risks letting misaligned data distort learning. Therefore, we propose a \textbf{Weighted} strategy that assigns higher weights to synthetic judgments that are more trustworthy, better aligned, and from underrepresented perspectives provides more reliable supervision. Each synthetic rating $x_i$, from persona $i$, receives a weight based on three components: 
(i) Alignment error: how closely $x_i$ matches human ratings; 
(ii) Perspective error: the trustworthiness of persona $i$’s demographic perspective (via k-means clustering on all demographic features \cite{vitsakis-etal-2024-voices} to identify intersectional identities, e.g., \textit{Black gen-z women with high school education});  and Perspective rarity: how underrepresented the demographic group is. 
The weight for each $x_i$ is:
\(
w_{x_i} = A(x_i) \cdot \frac{1}{T(c_i)} \cdot \frac{1}{P_{c_i}}
\). $A(x_i)$ is the \textit{alignment score}, inverse of the \textit{fidelity error}, the MAE between LLM-generated and empirical human ratings, averaged across demographic groups persona $i$ belongs to~\cite{simmons-savinov-2024-assessing}. $T(c_i)$ is the \textit{trustworthiness score} of persona $i$'s cluster $c_i$, based on MAE between model and real ratings. $P_{c_i}$ is the \textit{prevalence} of cluster $c_i$. Real data receive weights $w_{x_i} = 1$. During training, weights scale the loss for each synthetic rating:
\(
L = \sum_{i} w_{x_i} \cdot \text{Error}(y_i, \hat{y}_i)
\).
\camerareadytext{This strategy potentially gives more influence to synthetic ratings that are well-aligned, trustworthy, and demographically underrepresented.}




\subsection{Results}

We find that the impact of synthetic data on model performance is dataset specific, with no one approach consistently having positive impact (\fref{fig:exp3_all}). Our results somewhat support \textbf{(H3)}. Synthetic data is most helpful when it complements the underlying structure of disagreement: by enhancing diversity where consensus is weak and being applied sparingly in domains with high personalization. We summarize three observations.

\noindent \textbf{Datasets with high label disagreement benefit from blended training}. Offensiveness and Safety benefit the most from unweighted and weighted training strategies. Expanding the quantity (through unweighted training) or quality (through weighted training) of supervision might help to resolve low consensus. Surprisingly, adding randomly-selected synthetic annotations (0.5x and 1x) provided more gain than adding more demographically-curated ones (cluster).

\noindent \textbf{Dataset with high consensus see limited benefits}. Politeness (which has high global consensus but also high local disagreement) does not see any additional benefits from synthetic data. However, weighted training with cluster-generated synthetic data achieves the lowest MAE, as it could potentially resolve some local disagreements.


\noindent \textbf{Highly subjective domains do not benefit from synthetic data.} PCC, which is a highly subjective, complex, and personal annotation task that hinges on the annotator's personal experience with the healthcare system and assessment of interpersonal interactions, does not benefit from synthetic data. Cluster based generation method performs the worst, which may be due to the difficulty of clustering highly personal and idiosyncratic perceptions of doctor communication.

\section{Conclusion}
We present \name, a demographic-aware mixture of experts model that captures structured variation in annotator disagreement through group-level reasoning patterns. By routing inputs to expert subnetworks based on annotator demographics, \name introduces an inductive bias that improves performance on subjective judgments. Across three datasets, it outperforms other models in predicting ratings for nearly all demographic groups. To address sparse demographic coverage, we evaluate synthetic annotations from zero-shot LLM persona prompting and find moderate alignment with human ratings. We further blend real and synthetic annotations, showing that dataset-specific strategies enhance demographic alignment. These findings highlight annotator disagreement as a meaningful signal and offer practical methods for scaling perspective-aware learning in NLP.

\section{Limitations}
We find that our MoE model is most effective when the data exhibits high annotator disagreement, a strong demographic signal in the ratings, and sufficient annotation density. However, this finding may be limited by the fact that we apply our model to only five datasets, of which only three (PCC, Politeness, Offensiveness) contain sufficient demographic signals that the experts can leverage.

Additionally, we demonstrate the utility of MoE in modeling disagreement in tasks involving norm violations (e.g., Safety, Toxicity, Offensiveness, Politeness). However, it remains to be seen whether MoE can adapt effectively to disagreements in other domains, such as moral reasoning \cite{kumar2025rulesmeantbrokenunderstanding} or humor detection. Our approach also assumes that annotator disagreement reflects meaningful variation, though it may sometimes arise from noise or inconsistency. It would be valuable to assess the quality of disagreement -- such as by verifying annotator self-consistency or incorporating post-annotation deliberation -- to ensure that disagreements are substantive.

While we find that MoE experts tend to specialize in the perspectives of specific subgroups (e.g., expert 1 for Politeness focuses on the views of women and individuals with less than a high school education), our model structure does not explicitly represent intersectional identities. This lack of supervision may unintentionally essentialize identity. Future work could explore joint embeddings or hierarchical routing strategies (e.g., routing first by demographic category, then by intersectional identity).

By design, our MoE introduces many additional hyperparameters (e.g., weights in the loss function), beyond standard ones such as learning rate and batch size. Although we made extensive efforts to tune hyperparameters for all models, it is possible that we missed configurations that could improve their performance.

In the context of the PCC data (which we collected via Prolific), there may also be concerns about a biased sampling frame. Although we intentionally increased diversity by recruiting annotators to roughly balance key U.S. demographic groups (e.g., approximately 25\% White, 25\% Black, 25\% Asian, and 25\% Other; 50\% male, 50\% female) and collected detailed demographic information to assess coverage, the sample still reflects the characteristics of Prolific’s online participant pool. Such participants tend to be younger, more educated, and more technologically literate than the general population, and their experiences and expectations of healthcare communication may differ from those of typical patients. Consequently, the distribution of perspectives represented in our PCC annotations may not fully capture variation across less-represented or harder-to-reach populations. This limitation highlights the importance of validating models on datasets drawn from more representative or context-specific populations.

Next, the wording and formatting of annotation tasks could also influence our findings, both for the human annotations in PCC and the LLM-generated annotations in Experiment 2. In PCC, when asking annotators to rate dimensions of doctor communication (e.g., partnership), we frame the question around the concept underlying each construct (e.g., “encourages you to share your opinions”) rather than the construct label itself. This helps standardize annotators’ definitions and understanding based on the literature. However, providing only the construct name and relying on annotators’ folk understanding could yield different yet insightful results, potentially offering a closer simulation of how patients naturally interpret and evaluate physician communication in real encounters. For the LLM annotations, we kept prompts concise across tasks, but some models may benefit from more elaborated instructions. Systematically testing alternative wordings and formats could strengthen confidence in the LLM-generated annotations and improve downstream modeling using these synthetic data (Experiment 3).

Finally, the effectiveness of using synthetic data for training depends on both the quality of the data and the complexity of the task. While we experimented with different prompt lengths and wordings, there may be better configurations that enhance the fidelity of synthetic data. Our socio-demographic prompting (Experiment 2) could also benefit from techniques such as LoRA finetuning or few-shot learning \cite{orlikowski2025demographicsfinetuninglargelanguage}.

\section{Ethics}

Synthetic data offers a promising solution to the challenge of sparse demographic information, as it enables the scaling of diverse perspective modeling. However, using LLM-generated annotations for tasks such as PCC raises ethical concerns, as these ratings may reflect deeply personal and lived experiences shaped by the intersection of race, gender, and trust in the healthcare system. Simulating ratings based on sociodemographic inputs risks essentializing identities and producing stereotyped group profiles. Synthetic data may misrepresent or oversimplify the nuanced perspectives of minoritized groups. To mitigate this risk, we recommend that synthetic annotations be used sparingly in such tasks, and never as substitutes for real, diverse human judgments. Synthetic data should be clearly labeled, and its influence minimized through weighting based on its assessed trustworthiness. Even if a model shows strong performance across demographic groups, this may not equate to faithful or equitable representation of lived experiences—especially for marginalized populations.

A key downstream risk involves treating model outputs as ground truth. Because \name is trained to model group-level patterns from demographic data, its outputs may reflect aggregate tendencies rather than individual preferences—particularly for intersectional or underrepresented identities. Even with explicit model structuring, fairly representing intersectional identities remains a challenge due to the limited data available from minoritized groups. Training on such imbalanced datasets increases the risk of overfitting, which can introduce systemic biases. In practical applications, this poses significant implications. For example, if \name is trained on PCC data, it might be used to evaluate doctor communication in coaching contexts. Practitioners may mistakenly treat the model’s ratings as objective truth, without acknowledging that patients from different sociodemographic groups may experience the same interaction in markedly different ways. We therefore recommend treating model outputs as perspective-informed estimates, not universal judgments, and pairing them with real human input for proper context and interpretation.

\camerareadytext{\section*{Acknowledgments}

This work was supported in part by the National Science Foundation under Grant No. IIS-2143529.}

\bibliographystyle{acl_natbib}
\begingroup
\setlength{\parskip}{0pt}
\setlength{\itemsep}{0pt}
\setlength{\parsep}{0pt}
\setlength{\partopsep}{0pt}
\endgroup

\appendix

\section{Appendix}
\label{sec:appendix}

\subsection{Data}
\subsubsection{Data Overview}
\label{sec:data-overview}
We give an overview of the five datasets we use in Table. \ref{tab:dataset_stats}. 

\subsubsection{Demographic Signals}
\label{sec:demo-signal}
As a proxy for the strength of demgraphic signals, we report the ridge regression coefficients using the demographic group features for each dataset (Tables \ref{tab:safe_coef}, \ref{tab:tox_coef}, \ref{tab:pol_coef}, \ref{tab:off_coef}, \ref{tab:pcc_coef}). Safety shows the weakest demographic signal. 

\begin{table*}[t]
\centering
\rowcolors{2}{gray!8}{white} 

\resizebox{\textwidth}{!}{%
\begin{tabular}{@{}lrrrrrrrrp{3.5cm}p{5cm}@{}}
\toprule
\textbf{Dataset} & \textbf{\#Inst} & \textbf{\#Ann} & \textbf{\#Anns} & \textbf{\#Combos} & \textbf{Avg/Inst} & \textbf{IAA ($\alpha$)} & \textbf{Mean Entropy} & \textbf{Mean SD} & \textbf{Demographics} & \textbf{Task Description} \\
\midrule
Offensiveness \cite{pei2023annotatordemographicsmattermeasuring}& 1,500 & 262 & 25,042 & 177 & 8.69 & 0.287 & 1.212 & 0.909 & gender, race, age, occupation, education & Rate Reddit comment offensiveness (1--5). \\
Politeness \cite{pei2023annotatordemographicsmattermeasuring}   & 3,718 & 506 & 13,036 & 293 & 6.74 & 0.440 & 1.395 & 0.888 & gender, race, age, occupation, education & Rate the politeness of email (1--5). \\
Safety \cite{aroyo2023dicesdatasetdiversityconversational}       &  350  & 123 & 43,050 &  48 & 123.00 & 0.241 & 0.742 & 0.715 & gender, race, age, education & Rate harm in adversarial dialogue (1--3). \\
Patient Centered Communication & 2,230 & 589 & 7,553 & 478 & 3.33 & 0.287 & 1.492 & 0.849 & frequency of visiting healthcare providers in the last year, education, age, gender, race, occupation, level of trust toward doctors, level of ethnic-based trust toward medical system & Rating doctor qualities (informativeness, supportiveness, partnership) in doctor--patient conversations (1--5). Patient-centered communication is the average of these three. \\
Toxicity \cite{kumar2021designing} & 107,620 & 17,172 & 538,100 & 2,523 & 4.74 & 0.272 & 1.070 & 0.729 & gender, race, education, age range, political affiliation, LGBTQ status & Labeling the toxicity level of social media comments (1--5). \\
\bottomrule
\end{tabular}%
}
\caption{Dataset statistics. ``\#Inst'' = number of instances, ``\#Ann'' = annotators, ``\#Anns'' = annotations, ``\#Combos'' = unique demographic combinations, ``Avg/Inst'' = avg.\ annotators per instance, ``IAA'' = Krippendorff’s $\alpha$, ``Mean Entropy'' = average entropy per instance, and ``Mean SD'' = average of standard deviation of annotator ratings per instance. }
\label{tab:dataset_stats}
\end{table*}

\begin{table}[h!]
\centering
\caption{Ridge regression coefficients for Safety (sorted)}
\begin{tabular}{l c}
\hline
Feature & Coefficient \\
\hline
race & 0.5595 \\
age & 0.1248 \\
gender & 0.1125 \\
education & 0.1047 \\
\hline
\end{tabular}
\label{tab:safe_coef}
\end{table}

\begin{table}[h!]
\centering
\caption{Ridge regression coefficients for Toxicity (sorted)}
\begin{tabular}{l c}
\hline
Feature & Coefficient \\
\hline
age\_range & 2.0564 \\
education & 1.1142 \\
race & 0.7610 \\
lgbtq\_status & 0.6686 \\
gender & 0.5123 \\
political\_affiliation & 0.4758 \\
\hline
\end{tabular}
\label{tab:tox_coef}
\end{table}

\begin{table}[h!]
\centering
\caption{Ridge regression coefficients for Politeness (sorted)}
\begin{tabular}{l c}
\hline
Feature & Coefficient \\
\hline
race & 1.4273 \\
education & 0.8380 \\
age & 0.7597 \\
occupation & 0.6140 \\
gender & 0.2023 \\
\hline
\end{tabular}
\label{tab:pol_coef}
\end{table}

\begin{table}[h!]
\centering
\caption{Ridge regression coefficients for Offensiveness (sorted)}
\begin{tabular}{l c}
\hline
Feature & Coefficient \\
\hline
age & 1.3508 \\
occupation & 1.0809 \\
race & 0.9350 \\
education & 0.3848 \\
gender & 0.2556 \\
\hline
\end{tabular}
\label{tab:off_coef}
\end{table}

\begin{table}[h!]
\centering
\caption{Ridge regression coefficients for PCC (sorted)}
\begin{tabular}{l c}
\hline
Feature & Coefficient \\
\hline
age\_group & 2.5280 \\
gender & 1.5378 \\
edu\_level & 1.0890 \\
race & 0.8457 \\
doc\_trust\_category & 0.6098 \\
occupation & 0.5888 \\
hcp\_freq & 0.3029 \\
ethnic\_trust\_category & 0.2986 \\
\hline
\end{tabular}
\label{tab:pcc_coef}
\end{table}

\subsubsection{Patient Centered Communication Data}
\label{sec:pcc-details}

Our data for Patient Center Communication comes from transcripts of doctor-patient conversations during the PAACT (Partnering Around Cancer Clinical Trials) study \citep{Eggly2017}, whose goal was to test a multilevel intervention to increase the rates at which African-American and White men with prostate cancer make informed decisions to participate in a clinical trial. There were interventions for both physicians and patients: Physician intervention in communication include clinical communication (patient-centeredness, shared decision-making, consent), and relational communication (ask-tell-ask, lay language, teach-back); Patient intervention includes instructions and a list of questions related to clinical trials to encourage patients to participate actively. While the data was intentionally shared with us without personally identifiable information, its contents are nonetheless sensitive and the data use agreement prohibits resharing the data further---though the data remains available upon request. The original data was allowed for use and annotation with IRB approval \emph{anonymized number}.

Specifically, our data consists of 71 doctor-patient conversation transcripts on discussions between doctors and patients about prostate cancer treatment and trial enrollment. A summary table of the transcript is shown in Table \ref{tab:summary_stats}.  In addition to the transcript of the conversations, we also have access to patient sociodemographic information, and perception ratings (such as trust in a physician, and perceived physician patient-centered communication). There are also doctor measures (sociodemographic characteristics, attitudes toward clinical trials, implicit racial attitudes, etc). All measures are at multiple times during the trial (before the clinic visit, during the clinic visit, and in follow-up interview). 

\begin{table}[]
\begin{tabular}{|l|l|}
\hline
\begin{tabular}[c]{@{}l@{}}Total number of\\ conversations\end{tabular}                      & 71            \\ \hline
Total unique patients                                                                        & 51            \\ \hline
Total unique doctors                                                                         & 14            \\ \hline
\% of Black patients                                                                         & 46\%          \\ \hline
\% of White patients                                                                         & 54\%          \\ \hline
Average meeting time                                                                         & 20.54 minutes \\ \hline
\begin{tabular}[c]{@{}l@{}}Average total doctor \\ wordcount in a conversation\end{tabular}  & 1897.52 words \\ \hline
\begin{tabular}[c]{@{}l@{}}Average total patient \\ wordcount in a conversation\end{tabular} & 765.18 words  \\ \hline
\end{tabular}
 
 \caption{Summary statistics of PAACT transcript data}
 \label{tab:summary_stats}
\end{table}

\paragraph{Annotating PCC} We record doctor quality ratings of short conversation snippets with various measures collected in the original PAACT study, in addition to other well-studied measurements of patient perceptions of doctor qualities. The nine dimensions that we measure are: doctor partnership \cite{Street2007}, support \cite{Street2007}, informativeness \cite{Street2007}, warmth \cite{Howe2019}, empathy \cite{Sinclair2017}, respect \cite{Beach2006}; and patient perception of doctor's view on their communication \cite{Street2007}, agency \cite{agency_def}, and competence \cite{Ganzini2004} (e.g., ``to what extent does the doctor think that you are a good communicator?;;). To sample relevant snippets, we consider two criteria: 1) in the snippet, the doctor does not say too much backchannels; 2) the snippet should include enough context. Thus, we removed a snippet if the wordcount of doctor utterance is less than 25th percentile (excluding backchannel words); and if the doctor is the first speaker in the snippet, we included what the other person says right before the doctor. We kept each snippet to be 12 turns long. To augment the number of samples, we also slide the sampling window 6 turns after, resulting in a total of 2,232 snippets. 

We recruited 594 untrained annotators from the United States on Prolific. We aimed to increase the diversity of our own annotation data by sampling annotators on Prolific in a way that balanced key U.S. demographic groups (e.g., targeting approximately 25\% White, 25\% Black, 25\% Asian, and 25\% Other, and 50\% male, 50\% female). Instead of asking the participants directly about the measures, we reworded each to ensure precise definitions (Table \ref{tab:pcc-def}). We show each snippet to 4 different annotators to capture a variety of opinions. Annotators are shown 15 snippets of different conversations. They are asked to imagine that they are the patient in each snippet, and rate these dimensions of doctor qualities based on what the doctor says in each snippet. After completing the ratings, the annotators are also asked questions about their demographic information, their experience with the medical system, their trust in doctors, and ethnic group-based mistrust. Inter-annotator agreement as measured by Krippendorff's $\alpha$  ranges from 0.244 to 0.338 depending on the quality dimension (Section \ref{tab:iaa}), with doctor informativeness being the lowest, and doctor warmth being the highest. Cronbach's $\alpha$=0.958, meaning that although there is a lack of consensus among raters (as perceptions of doctor qualities are highly subjective depending on various factors such as experience with the medical system, or demographic factors), there is high internal consistency---i.e., annotators are likely to consistently give similar scores to similar questions about the same text. We aggregate the nine measurements into three measurements of doctor quality: 1) doctor patient-centered communication (sum of doctor informativeness, supportiveness, and partnership) \cite{Street2007} ; 2) doctor perception of patient communication (sum of patient communication, patient agency, and patient competence); and 3) doctor-patient relational communication (sum of doctor warmth, respect, and empathy) \cite{Hovey_Massfeller_2014, Back2005}.

\begin{table}[ht]
\centering
\small
\begin{tabularx}{\columnwidth}{l X}
\toprule
\textbf{Quality} & \textbf{Description} \\
\midrule

partnership   & encourages you to share your opinions \\
supportive   & is supportive of you \\
informative  & gives thorough and clear information \\
warmth       & is warm or kind towards you \\
empathy      & is empathetic towards you \\
respect      & is respectful towards you \\
communication & thinks you are engaged in the conversation and are communicating your preferences \\
agency       & thinks you can contribute to the conversation and decision-making \\
competence   & thinks you understand the situation \\
\bottomrule
\end{tabularx}
\caption{Definitions of patient-centered communication qualities.}
\label{tab:pcc-def}
\end{table}

 \begin{table}[tb]
\begin{tabular}{|l|l|}
\hline
\textbf{Measurement}   & \textbf{Krippendorff's alpha} \\ \hline
doctor informativeness & 0.2443                        \\ \hline
doctor partnership     & 0.2898                        \\ \hline
patient agency         & 0.2983                        \\ \hline
patient communication  & 0.3072                        \\ \hline
patient competence     & 0.3102                        \\ \hline
doctor respect         & 0.3263                        \\ \hline
doctor support         & 0.3279                        \\ \hline
doctor empathy         & 0.3354                        \\ \hline
doctor warmth          & 0.3380                        \\ \hline
\end{tabular}
\caption{Inter-annotator agreement for different ratings of the PAACT data.}
\label{tab:iaa}
\end{table}
\paragraph{Instructions Given to Annotators.}

\noindent \textit{\textbf{I. Consent}}
During this study, you will be asked to read 15 snippets of doctor-patient conversations, and then rate various doctor qualities. This survey is expected to take around 25 minutes. You will be compensated \$15.87/hr if you complete the survey. We cannot compensate you or use your data in our responses are of poor quality or if we find that your responses indicate you did not pay attention (e.g. nonsensical answers, continuous repetition of the same answers, lines copied and pasted form internet sources or AI, or impossibly low survey completion time).

The responses you provide will be used for research purposes only, specifically to train and evaluate models that predict how annotators rate doctor-patient communication. The models developed in this study will not be deployed in real-world systems at this stage and are intended solely for analysis, publication, and further academic research. There are no known risks to you from being in this research study. You are not expected to get any benefit from being in this research study. However, you may gain a better understanding of your attitudes and perceptions toward doctor-patient interactions. Additionally, your participation in this research study may benefit society by advancing our understanding of patient perceptions from various backgrounds. You can choose not to participate.

It is very important that you do not use AI to fill out any of the questions. Doing so will harm the quality of the data. Please answer these questions honestly. We are interested in getting diverse annotator perspectives.

Thank you for taking the time to participate in this research study!

If you have any questions about this study, feel free to contact the researcher below: [REDACTED]

By clicking the "I consent" choice below, you indicate that you have read the consent form.

You also understand that using AI to answer any of the survey questions means you will not be compensated.

\noindent \textit{\textbf{II. Instructions.}}

\begin{figure}[t]
    \centering
    \includegraphics[width=\columnwidth]{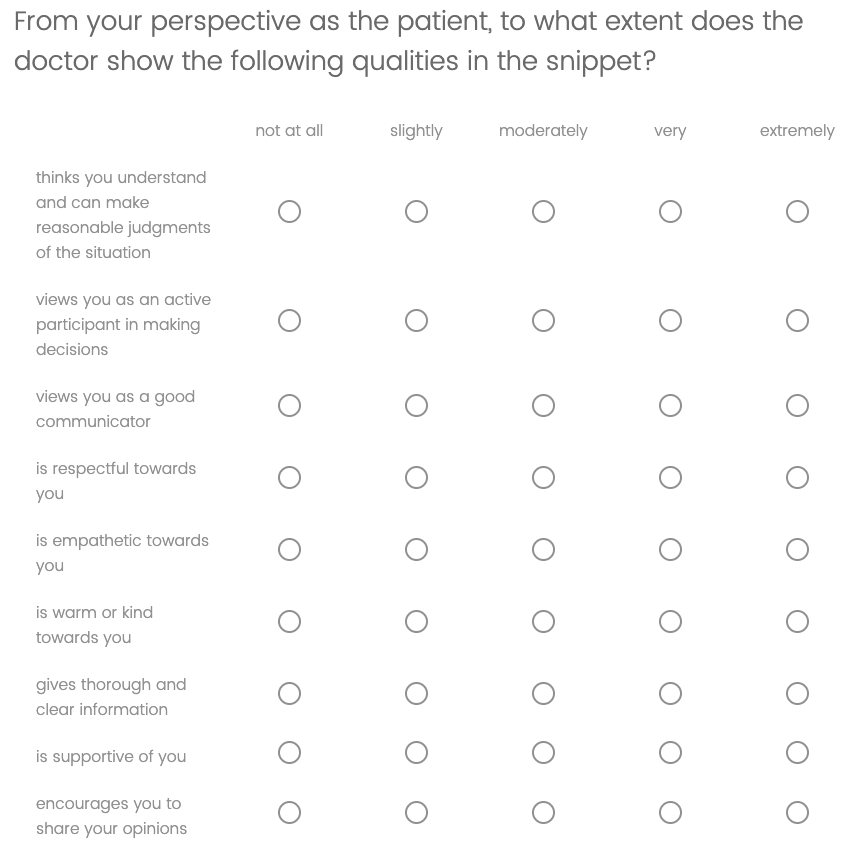}
    \caption{Screenshot of our questions.}
    \label{fig:screen-shot}
\end{figure}

This project aims to understand how people perceive doctor’s communication during their interactions with patients. You will see short snippets from various conversations between doctors and patients. You will be asked to rate how you feel about the doctor’s communication on several scales (e.g., respectfulness). In each conversation, the patient is diagnosed with prostate cancer and the doctor talks to him about his treatments. The doctor might talk about: the patient's health condition, a new trial or treatment, his eligibility to enroll in the trial, and the doctor’s recommendations. The conversation may include dialogue between doctors and family members/healthcare workers, but our focus is on the doctor.
Imagine you’re the patient in each snippet. From your perspective as the patient, you will rate the doctor’s qualities based on what the doctor says in each snippet. (For instance, based on the doctor’s behavior, do you think the doctor regards you, the patient, as a good communicator? ). You should rate based on the doctor’s general tone. In the rare case where you can’t judge one of the qualities, you can put “can’t tell”. Please rate these based on your understanding of the qualities. A screenshot of the questions are in Fig. \ref{fig:screen-shot}.

\paragraph{Annotator Demographics} We collected the following demographic attributes of annotators post-survey:

Annotator Past Experience Questions:
1) [hcp freq] During the past 12 months, not counting times you went to an emergency room, how many times did you go to a doctor, nurse, or other health professional to get care for yourself? \footnote{
\url{https://hints.cancer.gov/view-questions/question-detail.aspx?PK_Cycle=1&qid=711}}
\begin{itemize}
    \item None
    \item 1 time
    \item 2 times
    \item 3 times
    \item 4 times
    \item 5-9 times
    \item 10 or more times
\end{itemize}

2) [doc trust] Please rate the following from a scale from 1 - 5. Strongly Agree (5), Agree (4), Neutral (3), Disagree (2), Strongly Disagree (1).  \cite{Hall2002Trust}

\begin{itemize}
    \item Sometimes doctors care more about what is convenient for them than about their patients' medical needs.
    \item Doctors are extremely thorough and careful.
    \item You completely trust doctors' decisions about which medical treatments are best.
    \item A doctor would never mislead you about anything.
    \item All in all, you trust doctors completely.
\end{itemize}

3) [ethnic group-based trust]  Please rate the following on a scale of 1-5
 Strongly Agree (5), Agree (4), Neutral (3), Disagree (2), Strongly Disagree (1). \cite{THOMPSON2004209}
\begin{itemize}
    \item People of my ethnic group receive the same medical care from doctors and healthcare workers as people from other groups
    \item People of my ethnic group are treated the same as people of other groups by doctors and healthcare workers
    \item Doctors have the best interests of people of my ethnic group in mind
\end{itemize}

Annotator Demographic Questions: \footnote{https://hints.cancer.gov/docs/Instruments/HINTS6-AnnotatedEnglishInstrument.pdf}
\begin{enumerate}
    \item What is your gender identity?
    \item What is your current age?
    \item Which of the following best describe your current occupational status? Mark all the apply. (A) Employed. (B) Unemployed for 1 year or more. (C) Unemployed for less than 1 year. (D) Homemaker. (E) Student. (F) Retired. (G) Disabled. (H) Other
    \item What is the highest grade or level of schooling you completed? (A) Less than 8 years. (B) 8 through 11 years. (C) 12 years or completed high school. (D) Post high school training other than college (vocational or technical). (E) Some college. (F) College graduate. (G) Postgraduate.
    \item Are you of Hispanic or Latino origin or descent?
    \item What race or races do you consider yourself to be?
\end{enumerate}

\paragraph{Annotator Characteristics} The annotation survey resulted in 7553 total annotations. The top 10 most common annotator profiles are shown in Table \ref{tab:pcc-demo-profile}. The distributions for different subgroups in each demographic are shown in Table \ref{fig:pcc-dist}. For the purposes of modeling, we agregated some subcategories (e.g., hcp frequency originally had 7 categories, but we aggregated them to 3).

\begin{table*}[ht]
\centering
\small
\resizebox{\textwidth}{!}{%
\begin{tabular}{lllllllrl}
\toprule
\textbf{hcp\_freq} & \textbf{edu\_level} & \textbf{age\_group} & \textbf{gender} & \textbf{race} & \textbf{occupation} & \textbf{doc\_trust} & \textbf{ethnic\_trust} & \textbf{Count} \\
\midrule
3--9 times  & College Graduate or Higher          & 25 to 34 & Woman & Black & Employed & low trust           & low trust           & 52 \\
1--2 times  & College Graduate or Higher          & 25 to 34 & Woman & White & Employed & low trust           & low trust           & 45 \\
1--2 times  & College Graduate or Higher          & 45 to 64 & Man   & Black & Employed & high trust          & high trust          & 45 \\
1--2 times  & Some College or Vocational Training & 35 to 44 & Woman & White & Employed & moderate high trust & moderate high trust & 44 \\
1--2 times  & College Graduate or Higher          & 25 to 34 & Man   & Asian & Employed & high trust          & high trust          & 43 \\
1--2 times  & Some College or Vocational Training & 18 to 24 & Man   & White & Employed & moderate high trust & high trust          & 39 \\
1--2 times  & College Graduate or Higher          & 35 to 44 & Man   & Asian & Employed & high trust          & high trust          & 38 \\
1--2 times  & College Graduate or Higher          & 45 to 64 & Man   & White & Employed & high trust          & high trust          & 37 \\
1--2 times  & College Graduate or Higher          & 25 to 34 & Man   & Asian & Employed & moderate high trust & moderate high trust & 37 \\
1--2 times  & College Graduate or Higher          & 45 to 64 & Man   & Asian & Employed & high trust          & high trust          & 30 \\
\bottomrule
\end{tabular}}

\caption{[PCC] Top 10 most common demographic profiles.}
\label{tab:pcc-demo-profile}
\end{table*}

\section{\name Model Details and Training}
\label{app:train_loss}

\begin{figure*}[t]
    \centering
    \includegraphics[width=\textwidth]{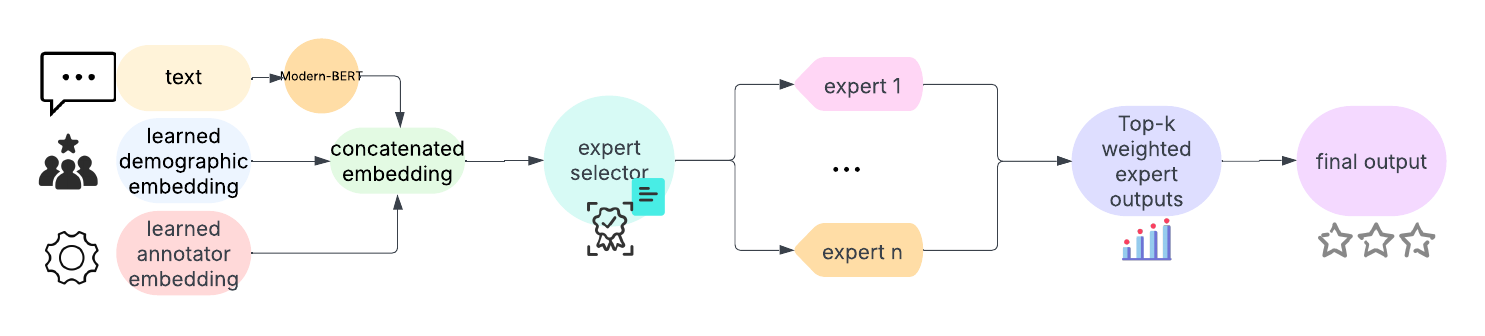}
    \caption{The architecture of our \name model}
    \label{fig:exp1_moe}
\end{figure*}

\name architecture is shown in Fig. \ref{fig:exp1_moe}.  We concatenate a text embedding, a learned annotator embedding, and learned demographic embedding and pass them through an expert selector, which produces logits over a shared pool of experts. At the expert selector, the inputs are directed to the most relevant experts, encoding the inductive bias. The top-$k$ experts are selected per sample, and their logits are normalized via softmax to generate weights. The final output, a single rating, is computed as a weighted combination of the top-$k$ expert outputs.

Our training loss  encodes inductive bias: 


\begin{align*}
\mathcal{L} =\; & 
\underbrace{\mathrm{MSE}(y, \hat{y})}_{\text{prediction}} 
+ \lambda_{\text{ann}}\, 
\underbrace{\mathrm{KL}\left(q(\mathbf{z}_a) \,\|\, \mathcal{N}(0, I)\right)}_{\text{annotator reg.}} \\
& + \lambda_{\text{id}}\, 
\underbrace{\mathrm{KL}\left(q(\mathbf{z}_g) \,\|\, \mathcal{N}(0, I)\right)}_{\text{demographic identity reg.}} \\
& + \lambda_{\text{load}}\, 
\underbrace{\mathrm{std}\left(\min\left(\frac{c_i}{\bar{c}}, 1\right)\right)}_{\text{load std}} \\
& + \lambda_{\text{orth}}\, 
\underbrace{\sum_{i,j \ne k} \frac{\langle x_{ij}, x_{ik} \rangle}{\langle x_{ik}, x_{ik} \rangle + \varepsilon}}_{\text{orthogonality}} \\
& + \lambda_{\text{var}}\, 
\underbrace{-\frac{1}{BE} \sum_{i=1}^B \sum_{j=1}^E (s_{ij} - \bar{s}_j)^2}_{\text{variance loss}} \\
& + \lambda_{\text{demo}}\, 
\underbrace{\sum_{d=1}^{D} \sum_{i < j} \frac{1}{2} \left[ \mathrm{KL}(p_i \| p_j) + \mathrm{KL}(p_j \| p_i) \right]}_{\text{demographic within-group specialization}}
\end{align*}

\begin{itemize}
  \item \(y, \hat{y}\): true and predicted scores
  \item \(q(\mathbf{z}_a)\), \(q(\mathbf{z}_g)\): posterior distributions for annotator and identity embeddings
  \item \(c_i\): token count routed to expert \(i\), \(\bar{c}\): average token count
  \item \(x_{ij}\): output from expert \(j\) for input \(i\)
  \item \(s_{ij}\): gating score for input \(i\), expert \(j\); \(\bar{s}_j\): expert \(j\)'s mean score
  \item \(p_i, p_j\): average expert distributions for demographic groups \(i, j\) within the same demographic (e.g., male vs female)
  \item \(D\): number of demographic attributes; \(B\): batch size; \(E\): number of experts
  \item \(\lambda_{\ast}\): task-specific hyperparameters for each loss component
\end{itemize}

\subsection{Experiment 1 Training Details}
\label{exp1op}

Jury learning models and ModernBERT are trained for 10 epochs, with early stopping. MoE models are trained for 50 epochs, with early stopping. During training, we tune the loss weights, in addition to learning rate. We find it helpful to apply the weights on load standard deviation, orthgonality loss, and variance loss in phases. Phase A has light penalties to encourage gating networks to start using multiple experts. Phase B has heavier penalties to ensure expert specialization. We keep the weights constant in Phase C to help stabilize the metrics. The transitions to different phases are determined by thresholds based on load standard deviation. For previously unseen annotators at test time, we assign the same default embedding that is randomly initialized once at model creation.

Using Optuna, we search hyperparameters with two iterations: we first start with the wider range of hyperparameter space, then narrow around the optimal hyperparameters. We use two different learning rates for the expert selector parameters vs. other parameters to ensure effective expert routing. We also gradually ramp up the load loss, orthogonal loss, and variance loss in different phases (A,B, and C). The thresholds for the phases are based on the expert load standard deviation. Phase A has light penalties to encourage gating networks to start using multiple experts. Phase B has heavier penalties to ensure expert specialization. We keep the weights constant in Phase C to help stabilize the metrics.

\subsubsection{Offensiveness}
We search the following hyperparameters for Offensiveness (Table \ref{tab:hyperparam_search_space_off}) to find the optimal values.

\begin{table}[ht]
\centering
\small
\resizebox{\columnwidth}{!}{%
\begin{tabular}{@{}llll@{}}
\toprule
\textbf{Hyperparameter} & \textbf{Search Range} & \textbf{Scale} & \textbf{Optimal Value} \\
\midrule

\texttt{learning\_rate\_gate}         & $[10^{-6}, 10^{-4}]$     & Log-uniform & 5.94e-5 \\
\texttt{learning\_rate\_main}         & $[5 \times 10^{-5}, 5 \times 10^{-3}]$ & Log-uniform & 1.58e-3 \\
\texttt{topk\_experts}        & $\{ 2,3\}$          & Discrete & 2 \\
\texttt{demographic\_emb\_w}  & $[10^{-6}, 10^{-3}]$     & Log-uniform & 0.0001 \\
\texttt{annotator\_emb\_w}    & $[10^{-5}, 10^{-2}]$     & Log-uniform & 0.001 \\
\texttt{demographic\_specialization\_w}    & $[0.15, 0.22]$     & Log-uniform & 0.0112 \\
\texttt{load\_loss\_w\_phaseA}       & $[0.1, 0.6]$             & Uniform & 0.261 \\
\texttt{load\_loss\_w\_phaseB}       & $[0.1, 0.6]$             & Uniform & 0.464 \\
\texttt{load\_loss\_w\_phaseC}       & $[0.3, 0.8]$             & Uniform & 0.897 \\
\texttt{orthogonal\_loss\_w\_phaseA}       & $[0.01, 0.2]$            & Uniform & 0.051 \\
\texttt{orthogonal\_loss\_w\_phaseB}       & $[0.1, 0.5]$             & Uniform & 0.252 \\
\texttt{orthogonal\_loss\_w\_phaseC}       & $[0.2, 0.6]$             & Uniform & 0.450 \\
\texttt{variance\_loss\_w\_phaseA}        & $[0.01, 0.2]$            & Uniform & 0.098 \\
\texttt{variance\_loss\_w\_phaseB}        & $[0.01, 0.2]$            & Uniform & 0.102 \\
\texttt{variance\_loss\_w\_phaseC}        & $[0.1, 0.5]$             & Uniform & 0.585 \\
\bottomrule
\end{tabular}}
\caption{Optuna hyperparameter search space and optimal values for key model parameters for Offensiveness.}
\label{tab:hyperparam_search_space_off}
\end{table}

\subsubsection{Politeness}
We search the following hyperparameters for Politeness (Table \ref{tab:hyperparam_search_space_pol}).

\begin{table}[ht]
\centering
\small
\resizebox{\columnwidth}{!}{%
\begin{tabular}{@{}llll@{}}
\toprule
\textbf{Hyperparameter} & \textbf{Search Range} & \textbf{Scale} & \textbf{Optimal Value} \\
\midrule
\texttt{learning\_rate\_gate}         & $[10^{-3}, 10^{-2}]$     & Log-uniform & 3.71e-3 \\
\texttt{learning\_rate\_main}         & $[10^{-3}, 10^{-2}]$     & Log-uniform & 3.78e-3 \\
\texttt{topk\_experts}                & $\{2, 3\}$               & Discrete    & 3 \\
\texttt{demographic\_emb\_w}          & $[10^{-4}, 10^{-2}]$     & Log-uniform & 7.09e-4 \\
\texttt{annotator\_emb\_w}            & $[10^{-4}, 10^{-2}]$     & Log-uniform & 5.35e-4 \\
\texttt{demographic\_specialization\_w} & $[0.05, 0.1]$          & Log-uniform & 0.0757 \\
\texttt{load\_loss\_w\_phaseA}        & $[0.2, 0.4]$             & Uniform     & 0.261 \\
\texttt{load\_loss\_w\_phaseB}        & $[0.4, 0.6]$             & Uniform     & 0.564 \\
\texttt{load\_loss\_w\_phaseC}        & $[0.7, 0.9]$             & Uniform     & 0.820 \\
\texttt{orthogonal\_loss\_w\_phaseA}  & $[0.01, 0.1]$            & Uniform     & 0.051 \\
\texttt{orthogonal\_loss\_w\_phaseB}  & $[0.2, 0.3]$             & Uniform     & 0.318 \\
\texttt{orthogonal\_loss\_w\_phaseC}  & $[0.4, 0.6]$             & Uniform     & 0.528 \\
\texttt{variance\_loss\_w\_phaseA}    & $[0.05, 0.15]$           & Uniform     & 0.098 \\
\texttt{variance\_loss\_w\_phaseB}    & $[0.15, 0.25]$           & Uniform     & 0.218 \\
\texttt{variance\_loss\_w\_phaseC}    & $[0.4, 0.6]$             & Uniform     & 0.467 \\
\bottomrule
\end{tabular}}
\caption{Optuna hyperparameter search space and optimal values for key model parameters for Politeness.}
\label{tab:hyperparam_search_space_pol}
\end{table}

\subsubsection{Safety}
We search the following hyperparameters for Safety (Table \ref{tab:hyperparam_search_space_dices}).

\begin{table}[ht]
\centering
\small
\resizebox{\columnwidth}{!}{%
\begin{tabular}{@{}llll@{}}
\toprule
\textbf{Hyperparameter} & \textbf{Search Range} & \textbf{Scale} & \textbf{Optimal Value} \\
\midrule
\texttt{learning\_rate\_gate}         & $[2 \times 10^{-4},\ 5 \times 10^{-4}]$     & Log-uniform & 3.00e-4 \\
\texttt{learning\_rate\_main}         & $[2 \times 10^{-4},\ 5 \times 10^{-4}]$     & Log-uniform & 3.07e-4 \\
\texttt{topk\_experts}                & $\{1,\ 2,\ 3\}$                             & Discrete    & 2 \\
\texttt{demographic\_emb\_w}          & $[5 \times 10^{-5},\ 2 \times 10^{-4}]$     & Log-uniform & 1.00e-4 \\
\texttt{annotator\_emb\_w}            & $[5 \times 10^{-4},\ 2 \times 10^{-3}]$     & Log-uniform & 1.00e-3 \\
\texttt{demographic\_specialization\_w} & $[0.15, 0.2]]$                                       & Log-uniform        & 0.186 \\
\texttt{load\_loss\_w\_phaseA}        & $[0.25,\ 0.3]$                             & Uniform     & 0.278 \\
\texttt{load\_loss\_w\_phaseB}        & $[0.5,\ 0.55]$                             & Uniform     & 0.528 \\
\texttt{load\_loss\_w\_phaseC}        & $[0.58,\ 0.63]$                            & Uniform     & 0.615 \\
\texttt{orthogonal\_loss\_w\_phaseA}  & $[0.08,\ 0.13]$                            & Uniform     & 0.108 \\
\texttt{orthogonal\_loss\_w\_phaseB}  & $[0.1,\ 0.15]$                             & Uniform     & 0.122 \\
\texttt{orthogonal\_loss\_w\_phaseC}  & $[0.5,\ 0.7]$                              & Uniform     & 0.652 \\
\texttt{variance\_loss\_w\_phaseA}    & $[0.15,\ 0.2]$                             & Uniform     & 0.172 \\
\texttt{variance\_loss\_w\_phaseB}    & $[0.12,\ 0.16]$                            & Uniform     & 0.143 \\
\texttt{variance\_loss\_w\_phaseC}    & $[0.3,\ 0.35]$                             & Uniform     & 0.348 \\
\bottomrule
\end{tabular}}
\caption{Optuna hyperparameter search space and optimal values for key model parameters for Offensiveness.}
\label{tab:hyperparam_search_space_dices}
\end{table}

\subsubsection{PCC}
We search the following hyperparameters for PCC (Table \ref{tab:hyperparam_search_space_pcc}).

\begin{table}[ht]
\centering
\small
\resizebox{\columnwidth}{!}{%
\begin{tabular}{@{}llll@{}}
\toprule
\textbf{Hyperparameter} & \textbf{Search Range} & \textbf{Scale} & \textbf{Optimal Value} \\
\midrule
\texttt{learning\_rate\_gate}         & $[1 \times 10^{-5},\ 2 \times 10^{-4}]$     & Log-uniform & 5.94e-5 \\
\texttt{learning\_rate\_main}         & $[1 \times 10^{-5},\ 2 \times 10^{-4}]$     & Log-uniform & 1.58e-3 \\
\texttt{topk\_experts}                & $\{2,\ 3,\ 4\}$                             & Discrete    & 3 \\
\texttt{demographic\_emb\_w}          & $[1 \times 10^{-5},\ 5 \times 10^{-4}]$     & Log-uniform & 1.37e-4 \\
\texttt{annotator\_emb\_w}            & $[5 \times 10^{-4},\ 0.01]$                 & Log-uniform & 1.17e-3 \\
\texttt{demographic\_specialization\_w} & $[0.01, 0.05]]$                                       & Log-uniform         & 0.0151 \\
\texttt{load\_loss\_w\_phaseA}        & $[0.05,\ 0.3]$                              & Uniform     & 0.130 \\
\texttt{load\_loss\_w\_phaseB}        & $[0.4,\ 0.7]$                               & Uniform     & 0.495 \\
\texttt{load\_loss\_w\_phaseC}        & $[0.6,\ 0.9]$                               & Uniform     & 0.745 \\
\texttt{orthogonal\_loss\_w\_phaseA}  & $[0.01,\ 0.1]$                              & Uniform     & 0.051 \\
\texttt{orthogonal\_loss\_w\_phaseB}  & $[0.2,\ 0.4]$                               & Uniform     & 0.256 \\
\texttt{orthogonal\_loss\_w\_phaseC}  & $[0.5,\ 0.8]$                               & Uniform     & 0.630 \\
\texttt{variance\_loss\_w\_phaseA}    & $[0.01,\ 0.1]$                              & Uniform     & 0.039 \\
\texttt{variance\_loss\_w\_phaseB}    & $[0.2,\ 0.4]$                               & Uniform     & 0.296 \\
\texttt{variance\_loss\_w\_phaseC}    & $[0.6,\ 0.9]$                               & Uniform     & 0.690 \\
\bottomrule
\end{tabular}}
\caption{Optuna hyperparameter search space and optimal values for key model parameters for PCC.}
\label{tab:hyperparam_search_space_pcc}
\end{table}

\subsubsection{Toxicity}
We search the following hyperparameters for Toxicity (Table \ref{tab:hyperparam_search_space_tox}).

\begin{table}[ht]
\centering
\small
\resizebox{\columnwidth}{!}{%
\begin{tabular}{@{}llll@{}}
\toprule
\textbf{Hyperparameter} & \textbf{Search Range} & \textbf{Scale} & \textbf{Optimal Value} \\
\midrule
\texttt{learning\_rate\_gate}         & $[1 \times 10^{-5},\ 2 \times 10^{-4}]$     & Log-uniform & 5.94e-5 \\
\texttt{learning\_rate\_main}         & $[5 \times 10^{-4},\ 0.003]$               & Log-uniform & 1.23e-3 \\
\texttt{topk\_experts}                & $\{2,\ 3\}$                                & Discrete    & 2 \\
\texttt{demographic\_emb\_w}          & $[1 \times 10^{-5},\ 2 \times 10^{-4}]$     & Log-uniform & 1.67e-4 \\
\texttt{annotator\_emb\_w}            & $[5 \times 10^{-4},\ 0.005]$                & Log-uniform & 1.41e-3 \\
\texttt{demographic\_specialization\_w} & $[0.5, 0.1]]$                                       & Log-uniform         & 0.0537 \\
\texttt{load\_loss\_w\_phaseA}        & $[0.3,\ 0.5]$                               & Uniform     & 0.401 \\
\texttt{load\_loss\_w\_phaseB}        & $[0.3,\ 0.6]$                               & Uniform     & 0.464 \\
\texttt{load\_loss\_w\_phaseC}        & $[0.4,\ 0.6]$                               & Uniform     & 0.520 \\
\texttt{orthogonal\_loss\_w\_phaseA}  & $[0.1,\ 0.3]$                               & Uniform     & 0.100 \\
\texttt{orthogonal\_loss\_w\_phaseB}  & $[0.1,\ 0.3]$                               & Uniform     & 0.124 \\
\texttt{orthogonal\_loss\_w\_phaseC}  & $[0.2,\ 0.4]$                               & Uniform     & 0.227 \\
\texttt{variance\_loss\_w\_phaseA}    & $[0.1,\ 0.3]$                               & Uniform     & 0.230 \\
\texttt{variance\_loss\_w\_phaseB}    & $[0.2,\ 0.4]$                               & Uniform     & 0.296 \\
\texttt{variance\_loss\_w\_phaseC}    & $[0.3,\ 0.5]$                               & Uniform     & 0.405 \\
\bottomrule
\end{tabular}}
\caption{Optuna hyperparameter search space and optimal values for key model parameters for Toxicity.}
\label{tab:hyperparam_search_space_tox}
\end{table}


\subsubsection{Training Details for Other Models}
We do grid search to find the optimal parameters. The optimal parameters for the Jury Learning models across all datasets are shown in Table \ref{tab:jury-params}. The optimal parameters for the Ea + En models across all datasets are in Table \ref{tab:uwua-params}. We used extra hyperparameters for finetuning on the Toxicity dataset because Jury Learning and Ea + En model due to their underperformance. We use optimal parameters for llama model following \cite{orlikowski2025demographicsfinetuninglargelanguage}.

\begin{table*}[t]
\small
\centering
\begin{adjustbox}{max width=\textwidth}
\begin{tabular}{@{}lcccccccccc@{}}
\toprule
\textbf{Task} &
\textbf{Cross Layers} &
\textbf{Dropout} &
\textbf{Batch Size} &
\textbf{MBERT LR} &
\textbf{CrossNet LR} &
\textbf{Demographic feedforward LR} &
\textbf{Regressor LR} &
\textbf{Optimizer LR} &
\textbf{Weight Decay} &
\textbf{Hidden Sizes} \\
\midrule
Toxicity &
5 & 0.3 & 256 &
5e-5 & 5e-4 & 5e-4 & 5e-4 & -- & 1e-4 &
128 / 256 \\

Safety &
5 & 0.2 & 16 &
-- & -- & -- & -- & 5e-6 & 1e-4 &
128 / 256 \\

Politeness &
5 & 0.2 & 32 &
-- & -- & -- & -- & 5e-5 & 1e-4 &
128 / 256 \\

Offensiveness &
5 & 0.2 & 8 &
-- & -- & -- & -- & 4e-6 & 1e-4 &
128 / 256 \\

PCC &
5 & 0.2 & 8 &
-- & -- & -- & -- & 4e-6 & 1e-4 &
128 / 256 \\
\bottomrule
\end{tabular}
\end{adjustbox}

\caption{Optimal hyperparameters for the Jury Learning model across all tasks. Dashes (--) indicate values not used for the task (e.g., MBERT-related LRs for tasks without frozen MBERT). Hidden sizes are shown as `embedding / feedforward`.}
\label{tab:jury-params}
\end{table*}

\begin{table*}[t]
\small
\centering
\begin{adjustbox}{max width=\textwidth}
\begin{tabular}{@{}lcc@{\hskip 6pt}cc@{\hskip 6pt}cc@{}}
\toprule
\textbf{Task} &
\textbf{Hidden Size} & \textbf{Dropout Rate} &
\textbf{Batch Size} & \textbf{Optimizer LR} &
\textbf{MBERT LR} & \textbf{Other Param LR} \\
\midrule
Toxicity     & 768  & 0.4 & 32  & --    & 2e-6 & 2e-5 \\
Politeness   & 1024 & 0.1 & 8   & 1e-6  & --   & --   \\
Offensiveness& 1024 & 0.1 & 100 & 1e-5  & --   & --   \\
PCC          & 1024 & 0.1 & 125 & 2e-5  & --   & --   \\
Safety       & 1024 & 0.1 & 32  & 2e-5  & --   & --   \\
\bottomrule
\end{tabular}
\end{adjustbox}
\caption{Optimal hyperparameters for the En + Ea model across all tasks. Dashes (--) indicate the parameter is not applicable for that task.}
\label{tab:uwua-params}
\end{table*}

\subsubsection{Computational Budget}
It takes around 20-30 minutes to run MoE models on PCC, Offensiveness, Politeness, and Safety on one NVIDIA RTX A6000 (Memory 48GB). It takes 1-2 hours to run MoE models on Toxicity. It takes about double the amount of time to run jury learning models. It takes about 2 hours to run Ea + En models on non-Toxicity datasets, and 4 hours to run on Toxicity.

\section{Additional Experiment 1 Results}
\label{app:more-exp1-results}

Here, we report additional experiments and results for Experiment 1.

\subsection{Performance on Seen vs. Unseen Annotators} 

We use three metrics: Pearson correlation ($r$), Mean Absolute Error (MAE), and Earth Mover's Distance (EMD). MAE and $r$ are calculated between predicted and actual annotator ratings, aggregated at the snippet level. $r$ measures how well the model captures directional alignment with human judgment, indicating consistency between predicted trends and actual data. MAE measures prediction accuracy and aligns with the primary metric in prior studies \cite{Gordon_2022}. EMD evaluates how well the model preserves opinion diversity by comparing predicted and true distributions of annotator ratings. \name, Jury Learning, LLaMA, and En + Ea model generate predictions at the annotator level, which we average to produce instance-level predictions. We then compute MAE and $r$ by comparing these to averaged annotator ratings per snippet. In contrast, ModernBERT does not model annotator-specific information and outputs instance-level ratings directly. To test model performance on seen vs unseen annotators, we use all three metrics for holistic evaluation.

\begin{figure}[htp]
    \centering    \includegraphics[width=0.9\columnwidth]{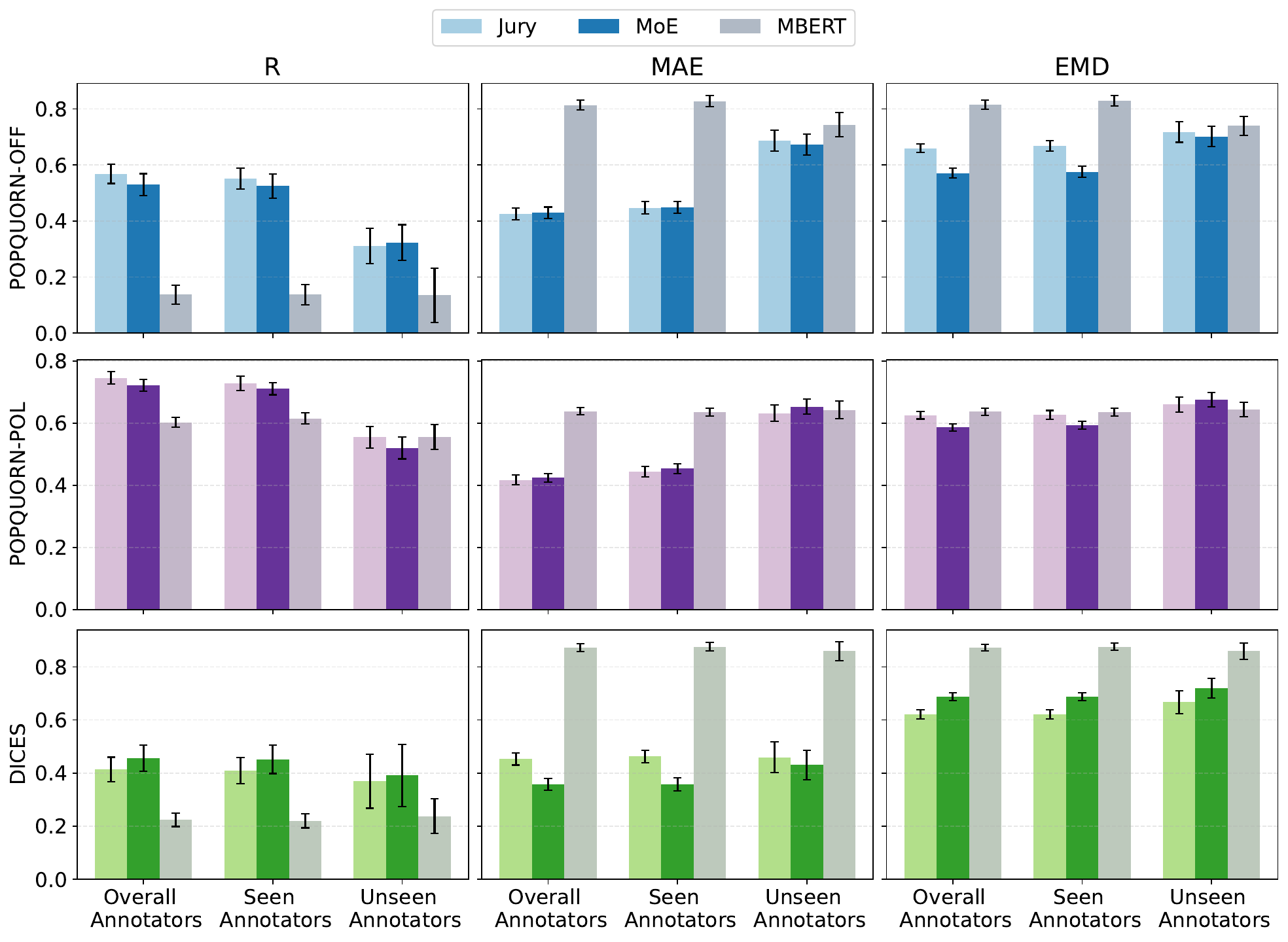}
    \caption{Model performance across datasets and metrics, for overall, annotators seen in the train set, and annotators not seen in the train set.}
    \label{fig:exp1_metrics}
\end{figure}

\name achieves comparable or superior performance to Jury Learning, MBERT, and PMF across all datasets and annotator groups (Fig. \ref{fig:exp1_metrics}). Gains are most notable on Safety, a most challenging dataset due to low inter-annotator agreement and diverse annotator pools. On Safety, MoE significantly outperforms Jury Learning in correlation and MAE, showing its ability to model complex, conflicting signals. On Offensiveness and PCC, MoE shows notable improvement in EMD, indicating better alignment with annotation distributions. On Politeness and Toxicity, MoE perform similarly as other SOTA models. These results suggest \name excels in low-agreement settings with dense annotator coverage.

Finally, MoE trains roughly twice as fast as Jury Learning. Its efficiency and strong representativeness make it well-suited for scenarios with large-scale, heterogeneous annotation data.

\subsection{Best zero-shot performance}
\label{app:best-zero}
Comparing these results (Table \ref{tab:offensiveness}, \ref{tab:politeness}, \ref{tab:safety}, \ref{tab:toxicity}, \ref{tab:pcc})  with the other models in Fig. 1, we see that zero-shot LLaMA consistently performs the worst across datasets because it is not optimized for modeling annotator judgments. Unlike PMF, it cannot capture systematic annotator behaviors or regularities, and unlike MBERT, it does not learn dataset-specific mappings from text features to annotator ratings. However, finetuning with LoRA (LoRA-LLaMA) substantially improves performance, bringing the model closer to the other baselines.

\begin{table}[ht]
\centering
\begin{tabularx}{\columnwidth}{lXXXX}
\hline
\textbf{Model} & \textbf{Zero-Shot Mean} & \textbf{Zero-Shot 95\% CI} & \textbf{LoRA Mean} & \textbf{LoRA 95\% CI} \\
\hline
Gender     & 0.847 & (0.842--0.853) & 0.708 & (0.690--0.726) \\
Race       & 0.918 & (0.904--0.931) & 0.565 & (0.510--0.619) \\
Age        & 0.872 & (0.865--0.879) & 0.757 & (0.740--0.775) \\
Occupation & 0.907 & (0.894--0.919) & 0.691 & (0.661--0.720) \\
Education  & 0.915 & (0.907--0.924) & 0.750 & (0.732--0.767) \\
\hline
\end{tabularx}
\caption{Best zero-shot performance on Offensiveness dataset.}
\label{tab:offensiveness}
\end{table}

\begin{table}[ht]
\centering
\begin{tabularx}{\columnwidth}{lXXXX}
\hline
\textbf{Model} & \textbf{Zero-Shot Mean} & \textbf{Zero-Shot 95\% CI} & \textbf{LoRA Mean} & \textbf{LoRA 95\% CI} \\
\hline
Gender     & 0.785 & (0.777--0.794) & 0.590 & (0.574--0.606) \\
Race       & 0.818 & (0.801--0.835) & 0.628 & (0.605--0.651) \\
Age        & 0.810 & (0.802--0.817) & 0.612 & (0.602--0.622) \\
Occupation & 0.794 & (0.788--0.801) & 0.590 & (0.580--0.600) \\
Education  & 0.773 & (0.765--0.781) & 0.585 & (0.574--0.596) \\
\hline
\end{tabularx}
\caption{Best zero-shot performance on Politeness dataset.}
\label{tab:politeness}
\end{table}

\begin{table}[ht]
\centering
\begin{tabularx}{\columnwidth}{lXXXX}
\hline
\textbf{Model} & \textbf{Zero-Shot Mean} & \textbf{Zero-Shot 95\% CI} & \textbf{LoRA Mean} & \textbf{LoRA 95\% CI} \\
\hline
Gender    & 0.874 & (0.869--0.879) & 0.627 & (0.613--0.641) \\
Race      & 0.890 & (0.884--0.898) & 0.644 & (0.628--0.660) \\
Age       & 0.881 & (0.875--0.886) & 0.628 & (0.622--0.634) \\
Education & 0.845 & (0.837--0.854) & 0.639 & (0.629--0.649) \\
\hline
\end{tabularx}
\caption{Best zero-shot performance on Safety dataset.}
\label{tab:safety}
\end{table}

\begin{table}[ht]
\centering
\begin{tabularx}{\columnwidth}{lXXXX}
\hline
\textbf{Model} & \textbf{Zero-Shot Mean} & \textbf{Zero-Shot 95\% CI} & \textbf{LoRA Mean} & \textbf{LoRA 95\% CI} \\
\hline
Gender                & 0.898 & (0.891--0.905) & 0.692 & (0.671--0.713) \\
Race                  & 0.943 & (0.940--0.946) & 0.804 & (0.788--0.820) \\
Education             & 0.910 & (0.907--0.913) & 0.749 & (0.737--0.761) \\
Age range             & 0.909 & (0.904--0.914) & 0.855 & (0.826--0.884) \\
Political affiliation & 0.921 & (0.920--0.922) & 0.745 & (0.725--0.765) \\
LGBTQ status          & 0.949 & (0.947--0.951) & 0.802 & (0.778--0.826) \\
\hline
\end{tabularx}
\caption{Best zero-shot performance on Toxicity dataset.}
\label{tab:toxicity}
\end{table}

\begin{table}[ht]
\centering
\begin{tabularx}{\columnwidth}{lXXXX}
\hline
\textbf{Model} & \textbf{Zero-Shot Mean} & \textbf{Zero-Shot 95\% CI} & \textbf{LoRA Mean} & \textbf{LoRA 95\% CI} \\
\hline
hcp freq     & 0.906 & (0.898--0.914) & 0.841 & (0.827--0.855) \\
edu level    & 0.980 & (0.967--0.992) & 0.919 & (0.882--0.956) \\
age group    & 0.933 & (0.924--0.943) & 0.839 & (0.827--0.851) \\
gender       & 0.968 & (0.937--1.000) & 0.899 & (0.846--0.952) \\
race         & 0.975 & (0.966--0.984) & 0.838 & (0.828--0.848) \\
occupation   & 0.979 & (0.967--0.992) & 0.916 & (0.892--0.940) \\
doc trust    & 0.955 & (0.947--0.963) & 0.856 & (0.842--0.870) \\
ethnic trust & 0.969 & (0.961--0.978) & 0.881 & (0.865--0.897) \\
\hline
\end{tabularx}
\caption{Best zero-shot performance on PCC dataset.}
\label{tab:pcc}
\end{table}
\FloatBarrier

\subsection{Capacity vs. Inductive Bias}
Model capacity could potentially confound MoE’s increased performance compared to other models. However, we don't see this to be the case based on the parameter counts (Table \ref{tab:param_counts}). We see that actually LoRA-LLaMA, MBERT, and En + Ea Model have much larger parameter sizes compared to MoE. However, despite having fewer parameters, MoE consistently outperforms other models in most demographic categories on Toxicity, Offensiveness, and PCC. Thus we can confidently say that our proposed architecture, rather than the increase in parameter capacity, provides the beneficial inductive bias. 

\begin{table}[ht]
\centering
\begin{tabular}{l r}
\hline
\textbf{Model} & \textbf{Trainable Parameters (M)} \\
\hline
PMF            & 0.22 \\
LoRA-LLaMA     & \textbf{3.43} \\
MBERT          & \textbf{394.78} \\
En + Ea Model  & \textbf{4.80} \\
Jury Learning  & 0.49 \\
MoE            & \textbf{2.47} \\
\hline
\end{tabular}
\caption{Trainable parameter counts for all models compared in the main results.}
\label{tab:param_counts}
\end{table}

\subsection{Experiment 1 within-group expert specialization}
\begin{figure}[htp]
    \centering    \includegraphics[width=0.9\columnwidth]{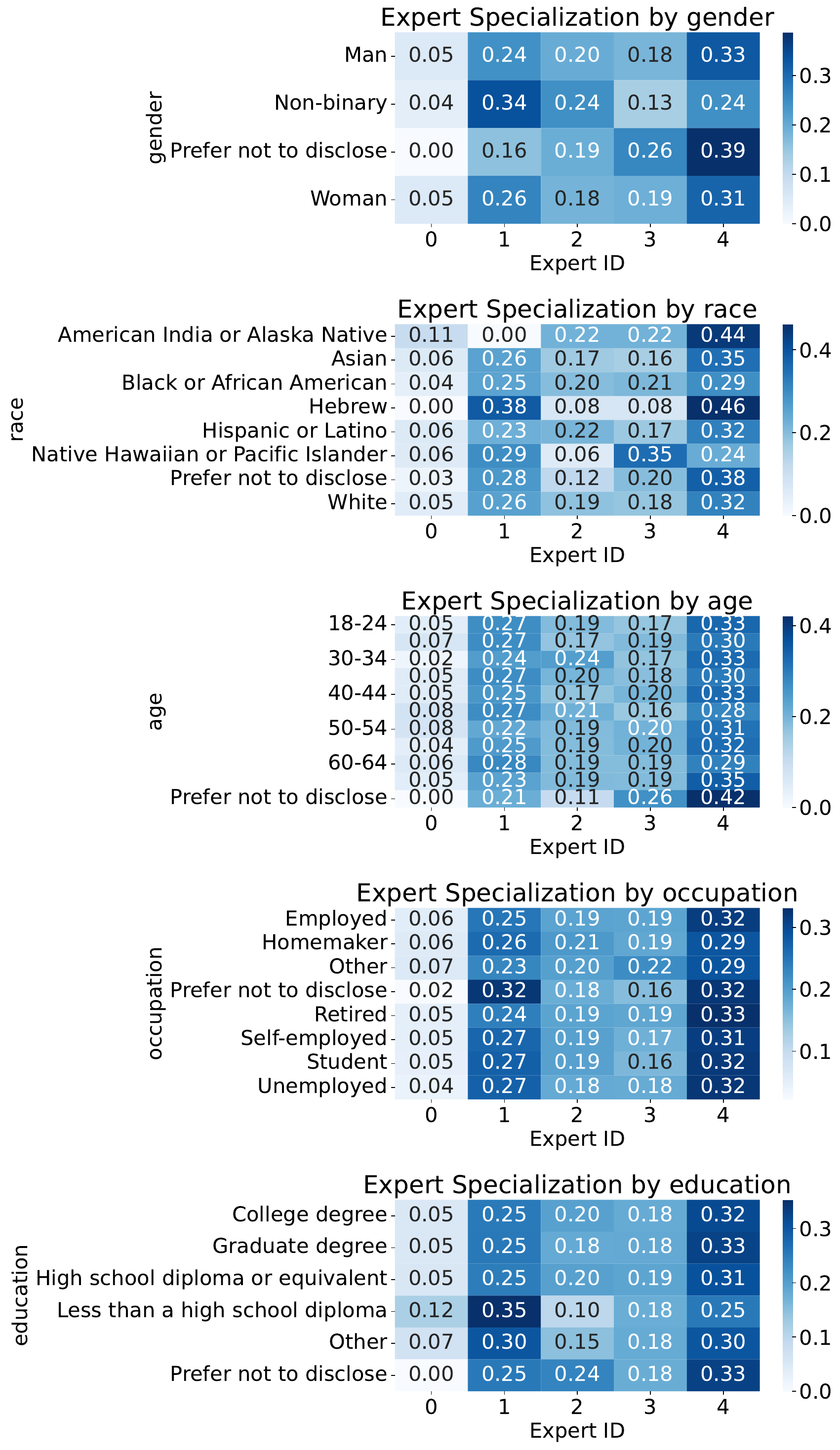}
    \caption{[Politeness] Expert usage for each demographic category, normalized by each subgroup (row).}
    \label{fig:pol-within-group}
\end{figure}

\begin{figure}[htp]
    \centering    \includegraphics[width=0.9\columnwidth]{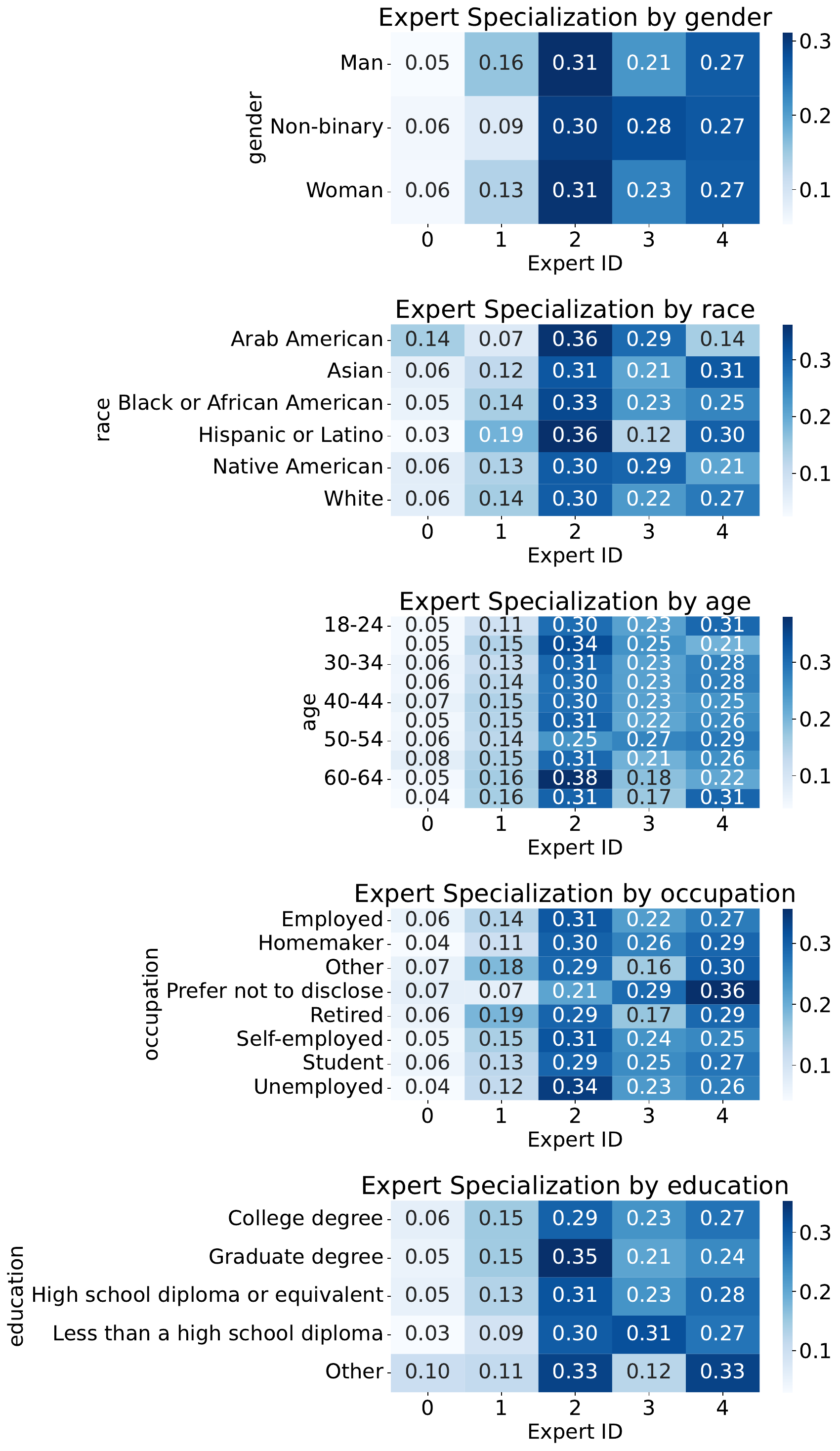}
    \caption{[Offensiveness] Expert usage for each demographic category, normalized by each subgroup (row).}
    \label{fig:off-within-group}
\end{figure}

\begin{figure}[htp]
    \centering    \includegraphics[width=0.9\columnwidth]{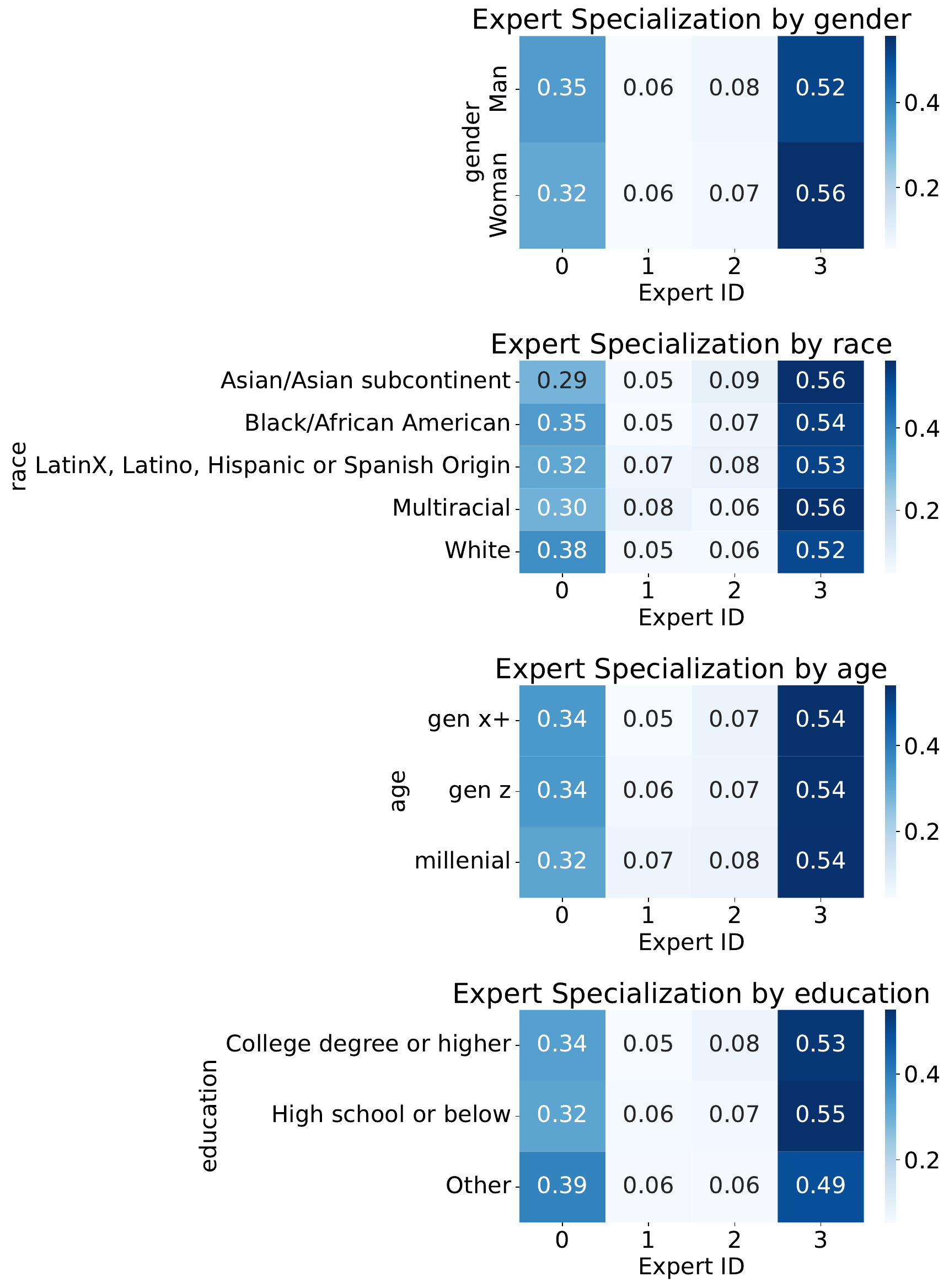}
    \caption{[Safety] Expert usage for each demographic category, normalized by each subgroup (row).}
    \label{fig:safe-within-group}
\end{figure}

\begin{figure}[htp]
    \centering    \includegraphics[width=0.9\columnwidth]{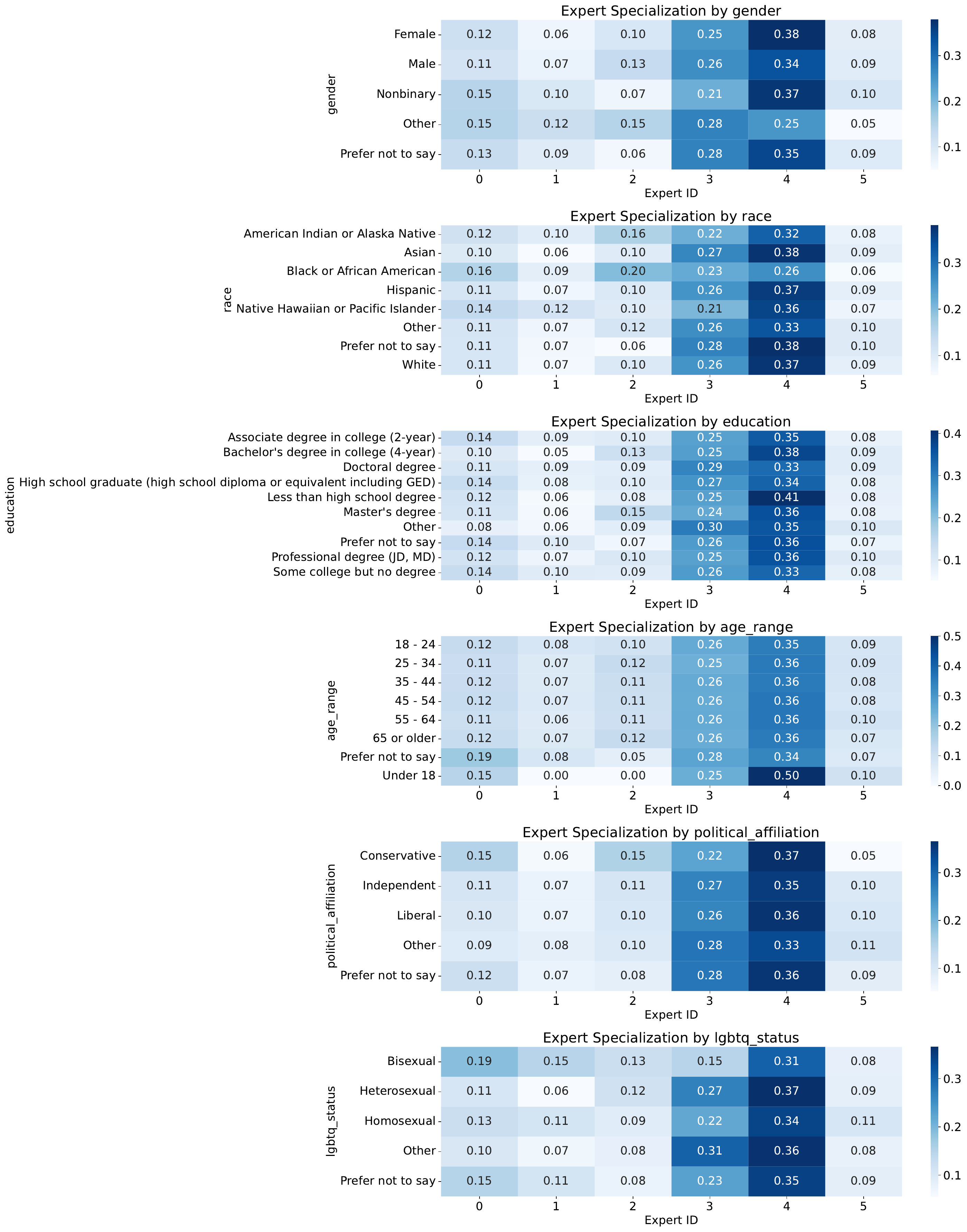}
    \caption{[Toxicity] Expert usage for each demographic category, normalized by each subgroup (row).}
    \label{fig:tox-within-group}
\end{figure}

\begin{figure}[htp]
    \centering    \includegraphics[width=0.9\columnwidth]{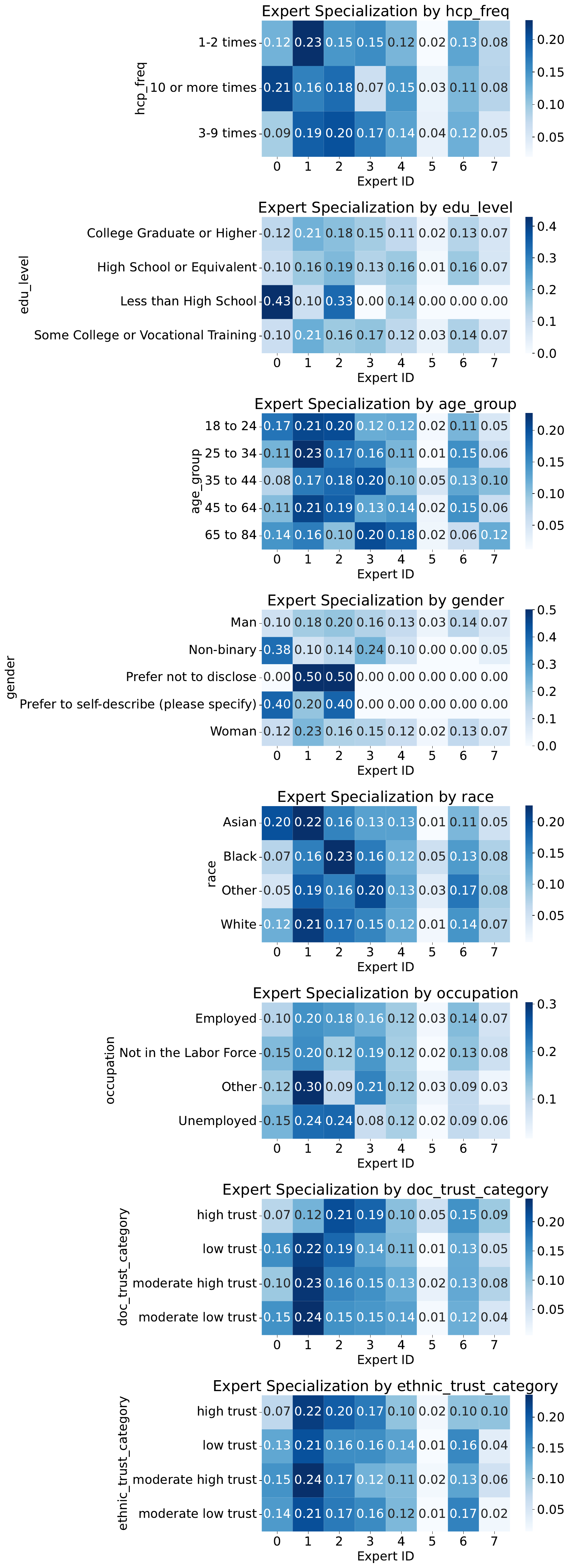}
    \caption{[PCC] Expert usage for each demographic category, normalized by each subgroup (row).}
    \label{fig:pcc-within-group}
\end{figure}
Expert usage for each demographic category is shown in Fig. \ref{fig:pol-within-group} for Politeness, Fig. \ref{fig:off-within-group} for Offensiveness, Fig. \ref{fig:safe-within-group} for Safety, Fig. \ref{fig:tox-within-group} for Toxicity, and Fig. \ref{fig:pcc-within-group} for PCC.  The figures show subgroup-level averages, where we aggregate expert usage across all instances within each demographic category. This averaging naturally produces distributions where multiple experts appear active, even though routing is discrete per instance.

For Politeness, we see that there is sufficient expert specialization: expert 1 specializes in the perspective of non-binary people, Hebrew, and people with an education less than a high school diploma; expert 4 specializes in prefer not to disclose (gender), Hebrew, Prefer not to disclose (age), and most of the perspectives in occupation and education. 

For offensiveness, Expert 2 specializes in the perception of all three gender categories, Arab and Latino American, adults ages 60-64, unemployed people, and people with a Graduate / other degree. Expert 3 specializes in perspectives from non-binary people, Native American, and people with less than a high school diploma.

For Safety, we see inadequate within group expert specialization: experts 0 and 3 primarily dominate in representing all perspectives, potentially due to the low predictive power of the demographic variables on annotation ratings.

For toxicity, we see better specialization. Most perspectives are specialized by experts 3 and 4, but expert 2 specializes in perspectives from African Americans, people with Master's degree, and conservatives.

For PAACT, we see sufficient with-in group expert specialization. For instance, expert 1 specializes in the perspective of annotators who are young to middle-aged, who rarely visit healthcare professionals, who are Asian and White, and who have low to moderate trust in the medical profession but high ethnic-based group trust in the medical system. On the other hand, expert 2 specializes in annotators who visit healthcare professionals a moderate number of times, people with less than high school education, younger annotators, Black annotators, and people with high and ethnic-based trust toward the medical system.

\subsection{Experiment 1 Cross-group expert specialization}

\begin{figure}[htp]
    \centering    \includegraphics[width=0.6\columnwidth]{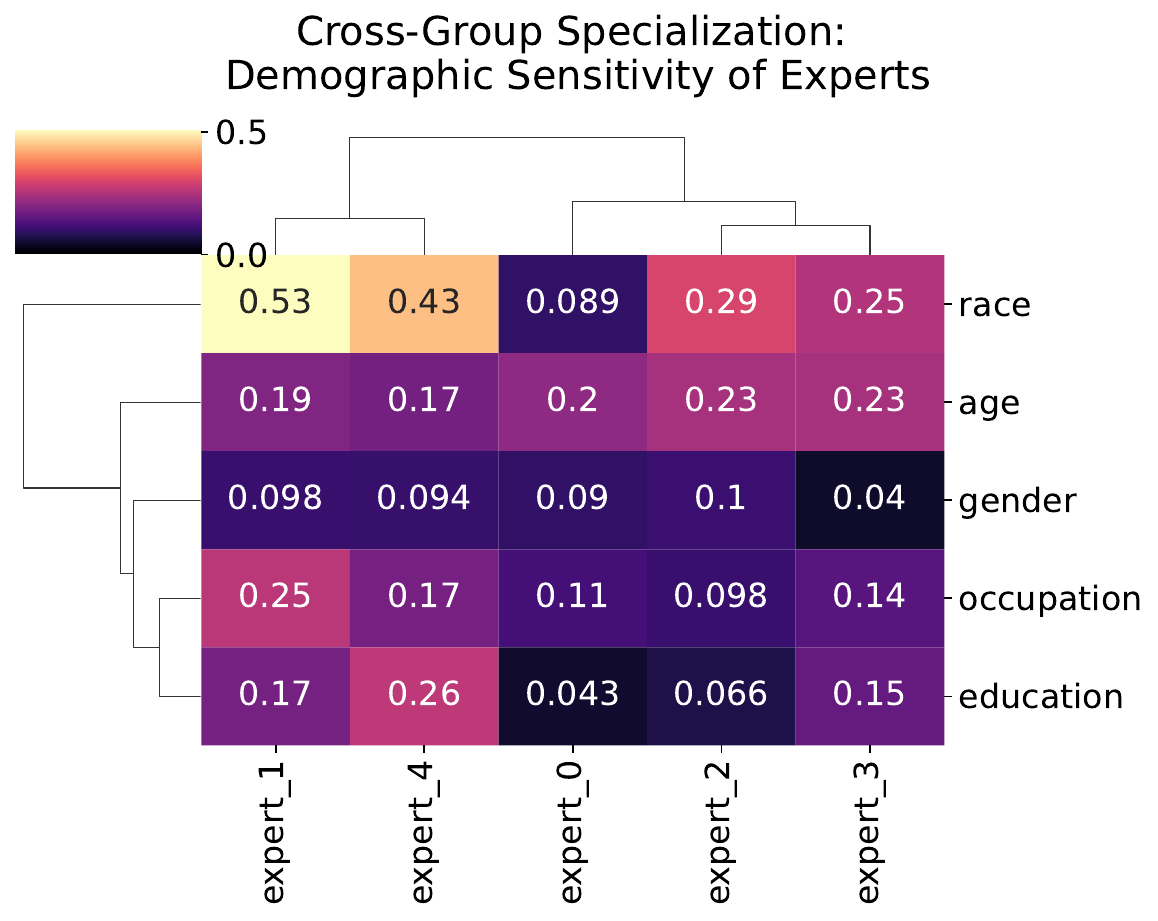}
    \caption{[Politeness] Clustered heatmap of ridge regression coefficients, where demographic attributes are used to predict expert usage.}
    \label{fig:pol-cross-group}
\end{figure}

\begin{figure}[htp]
    \centering    \includegraphics[width=0.6\columnwidth]{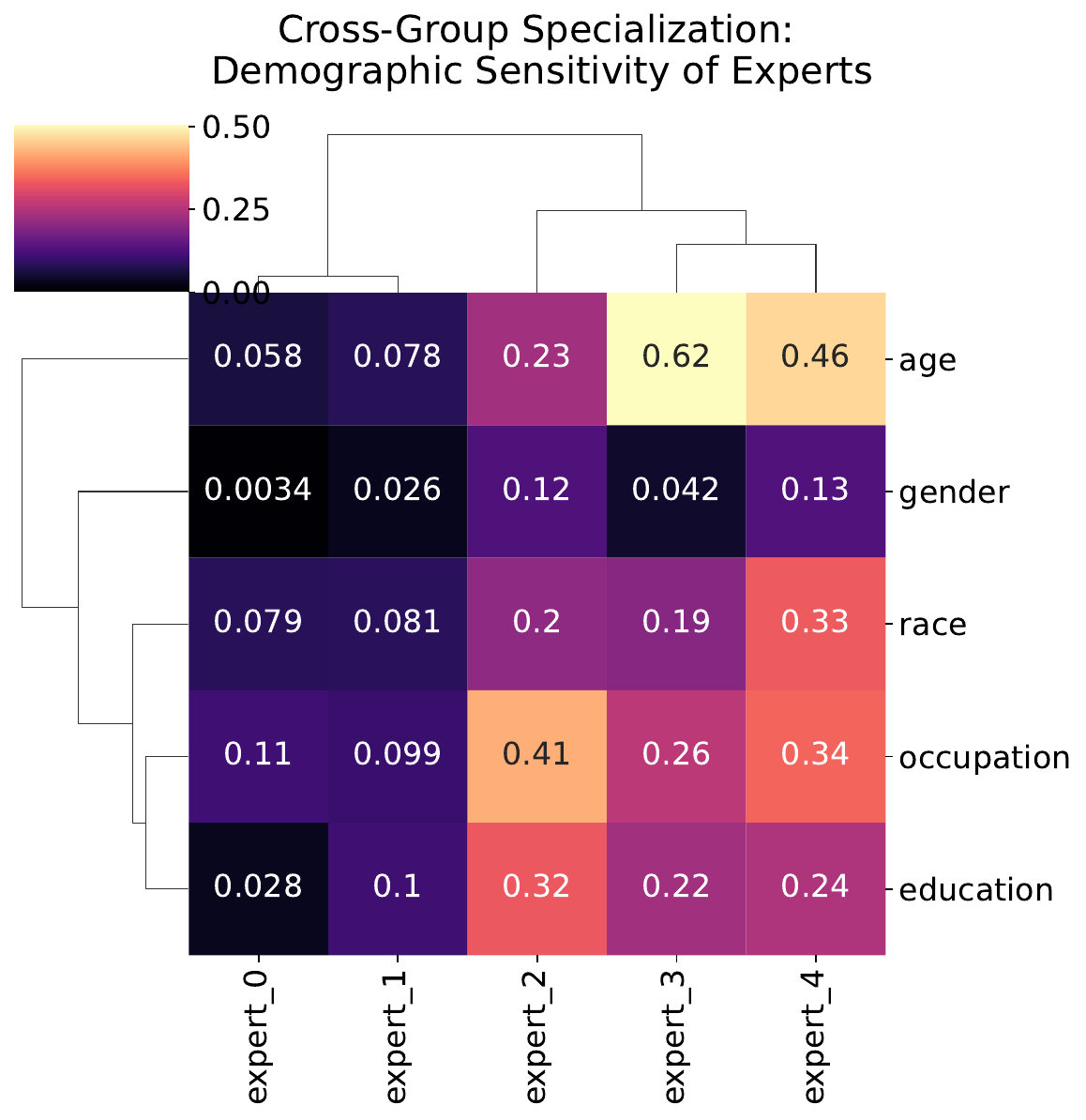}
    \caption{[Offensiveness] Clustered heatmap of ridge regression coefficients, where demographic attributes are used to predict expert usage.}
    \label{fig:off-cross-group}
\end{figure}

\begin{figure}[htp]
    \centering    \includegraphics[width=0.6\columnwidth]{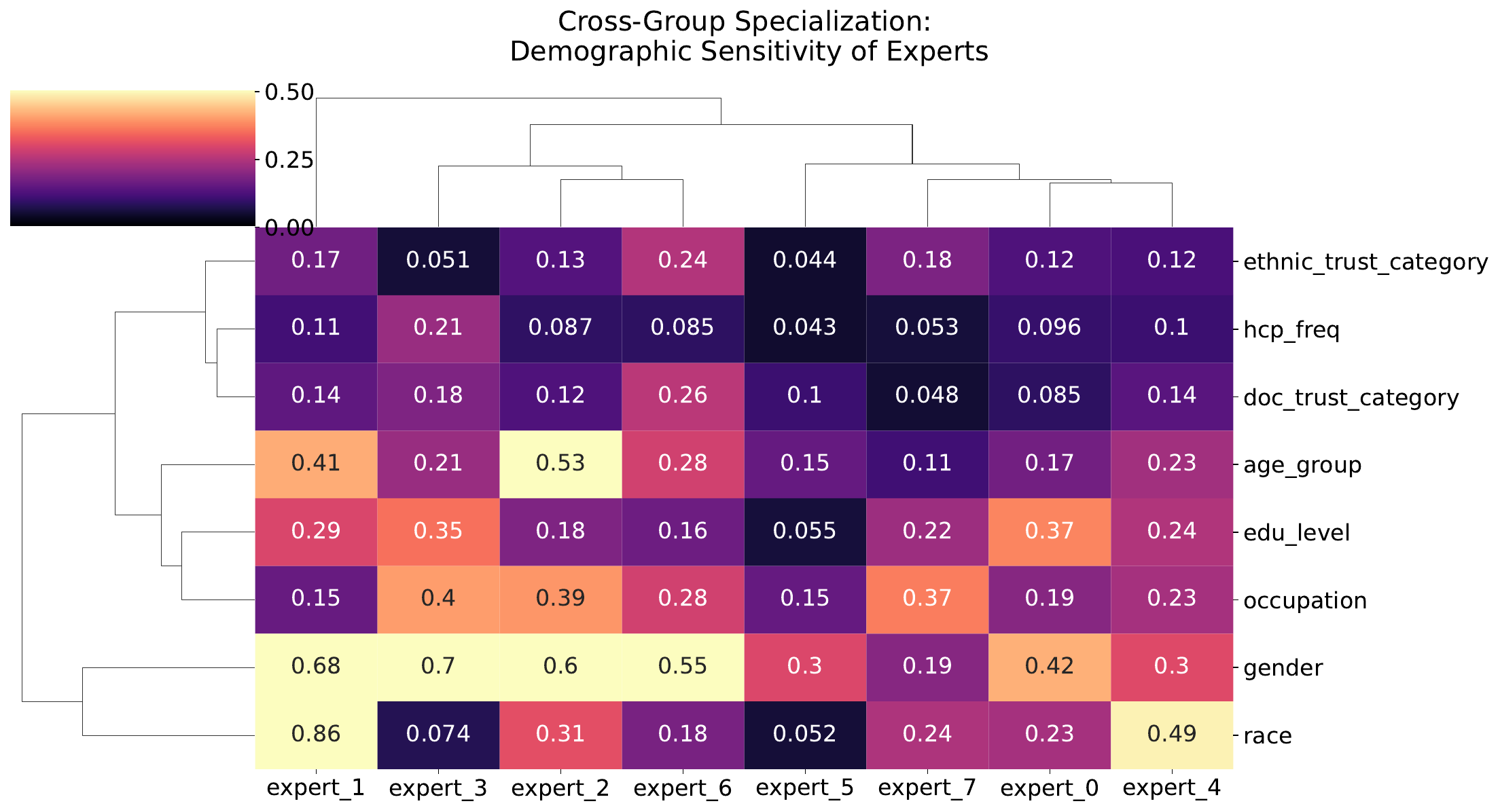}
    \caption{[PCC] Clustered heatmap of ridge regression coefficients, where demographic attributes are used to predict expert usage.}
    \label{fig:pcc-cross-group}
\end{figure}

\begin{figure}[htp]
    \centering    \includegraphics[width=0.6\columnwidth]{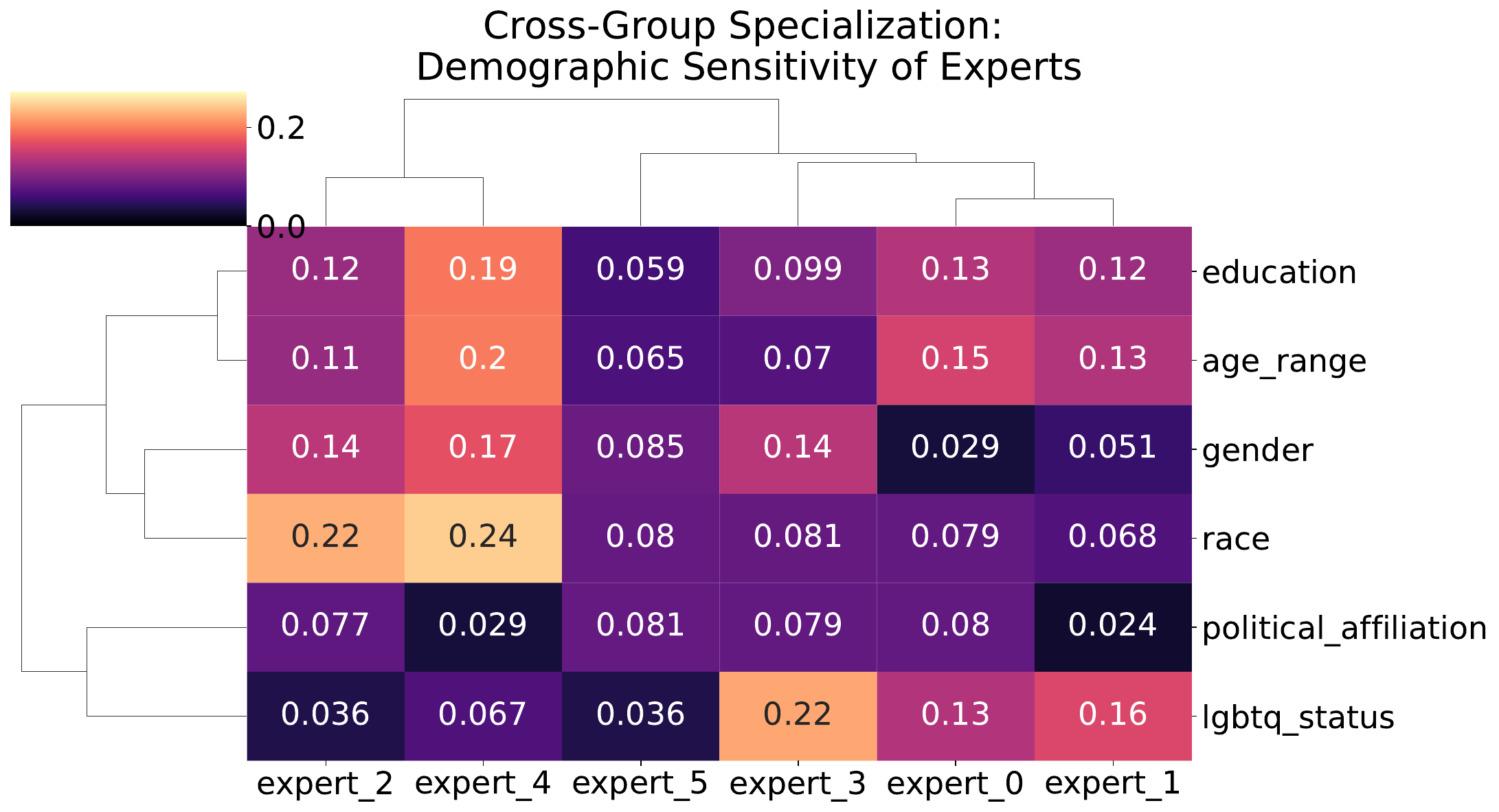}
    \caption{[Toxicity] Clustered heatmap of ridge regression coefficients, where demographic attributes are used to predict expert usage.}
    \label{fig:tox-cross-group}
\end{figure}

\begin{figure}[htp]
    \centering    \includegraphics[width=0.6\columnwidth]{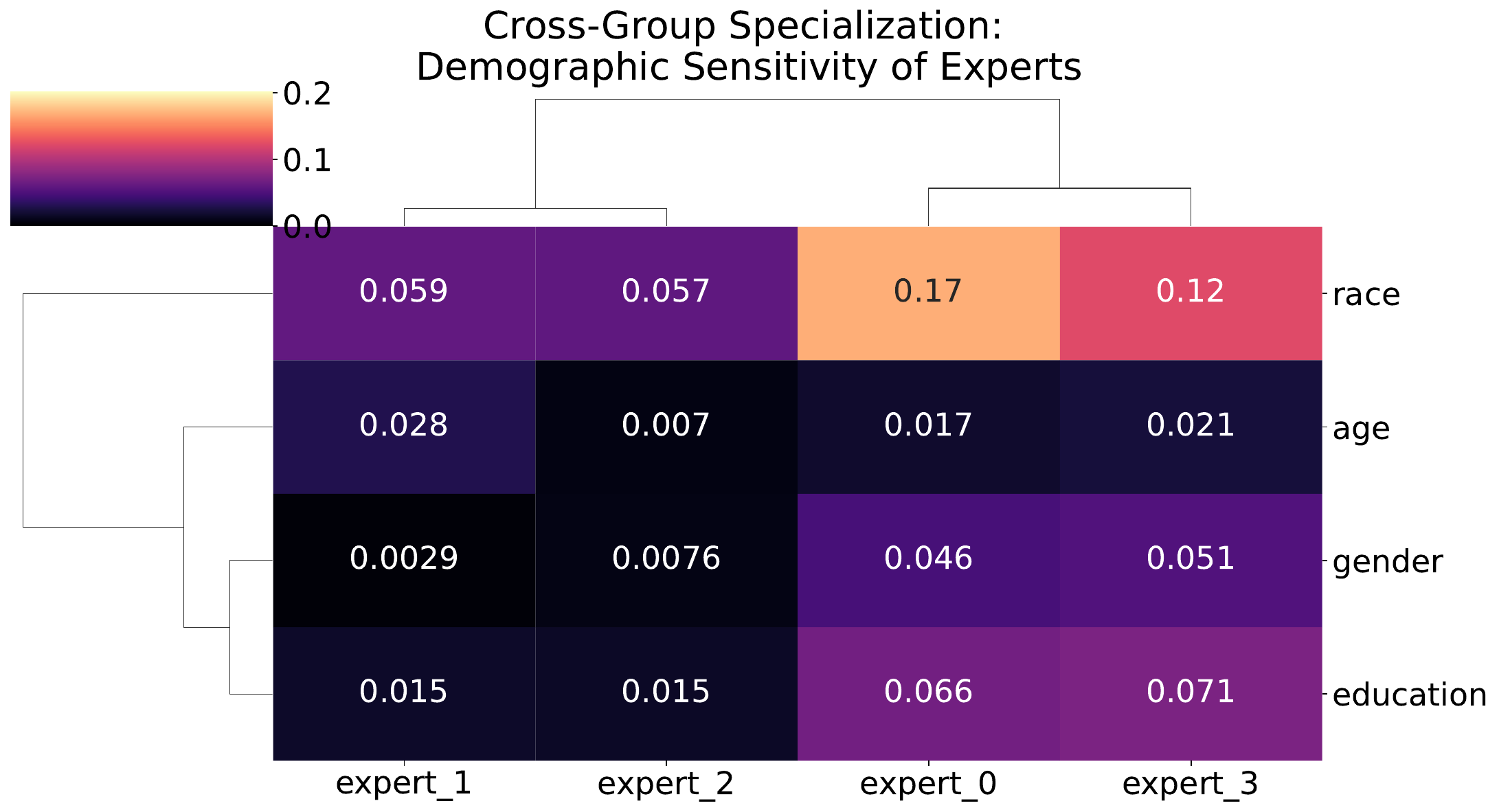}
    \caption{[Safety] Clustered heatmap of ridge regression coefficients, where demographic attributes are used to predict expert usage.}
    \label{fig:safety-cross-group}
\end{figure}
To analyze cross-group expert specialization, we use ridge regression to predict expert usage from demographic attributes, and visualize the coefficients in a clustered heatmap. For Politeness (Fig. \ref{fig:pol-cross-group}), expert 1 specializes in race and occupation, and expert 4 specializes in race and education. Both experts 2 and 3 specialize in race and age. 

For Offensiveness (Fig. \ref{fig:off-cross-group}), both experts 3 and 4 specialize in age and occupation, and expert 2 in occupation and education.

For PCC (Fig. \ref{fig:pcc-cross-group}), experts 1,2,3,6 all specialize in gender. Expert 2 and 6 are similar in that they also specialize in age group. Both experts 0 and 4 are similar in the sense that they specialize in race, but expert 0 also specializes in education level, and expert 4 specializes in race. 

For Toxicity (Fig. \ref{fig:tox-cross-group}), both experts 2 and 4 specialize in education, age, gender, and race. Both experts 0 and 1 specialize in education, age, and LGBTQ status. Expert 3 specializes in gender and LGBTQ status.

For Safety (Fig. \ref{fig:safety-cross-group}), both experts 0 and 3 specialize in race and, to a lesser extent, education and gender.

\section{Additional Experiment 2 Details and Results}
\label{app:more-exp2-results}

\subsection{Experiment 2 Group MAE Results}
\label{app:exp2-group-results}
\begin{figure}[htp]
    \centering    \includegraphics[width=0.9\columnwidth]{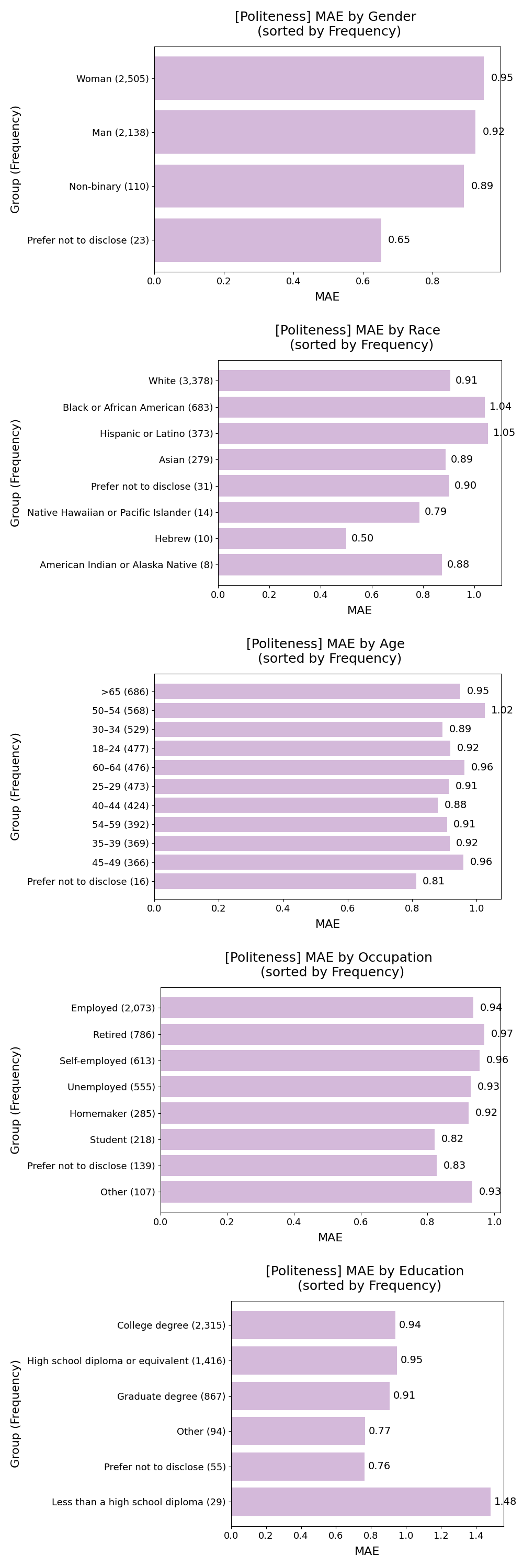}
    \caption{MAE by demographic group for the Politeness dataset.}
    \label{fig:pol_exp2_group}
\end{figure}

\begin{figure}[htp]
    \centering    \includegraphics[width=0.9\columnwidth]{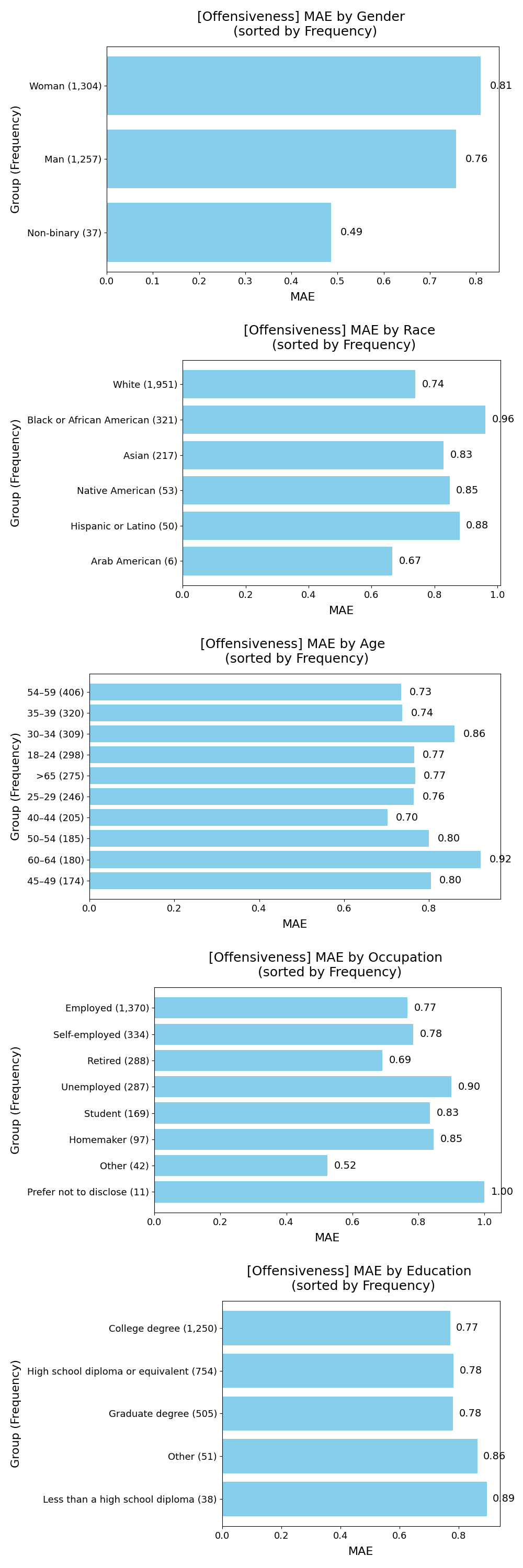}
    \caption{MAE by demographic group for the Offensiveness dataset.}
    \label{fig:off_exp2_group}
\end{figure}

\begin{figure}[htp]
    \centering    \includegraphics[width=0.9\columnwidth]{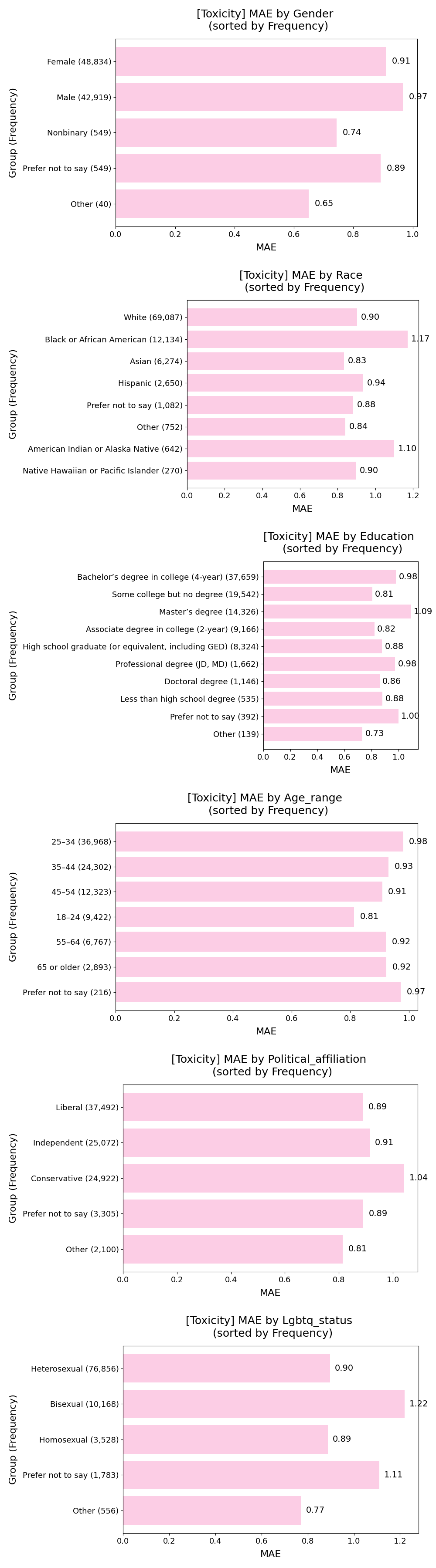}
    \caption{MAE by demographic group for the Toxicity dataset.}
    \label{fig:tox_exp2_group}
\end{figure}

\begin{figure}[htp]
    \centering    \includegraphics[width=0.9\columnwidth]{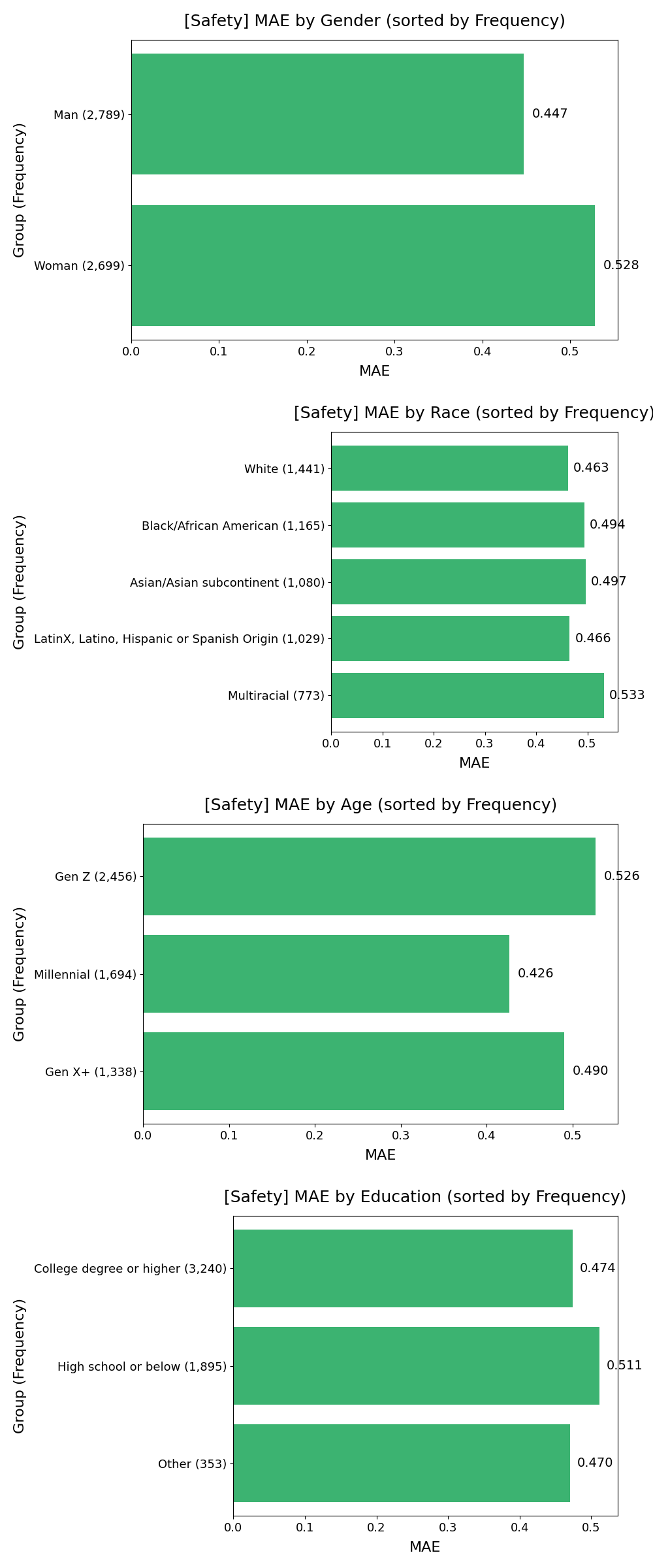}
    \caption{MAE by demographic group for the Safety dataset.}
    \label{fig:safety_exp2_group}
\end{figure}

\begin{figure}[htp]
    \centering    \includegraphics[width=0.6\columnwidth]{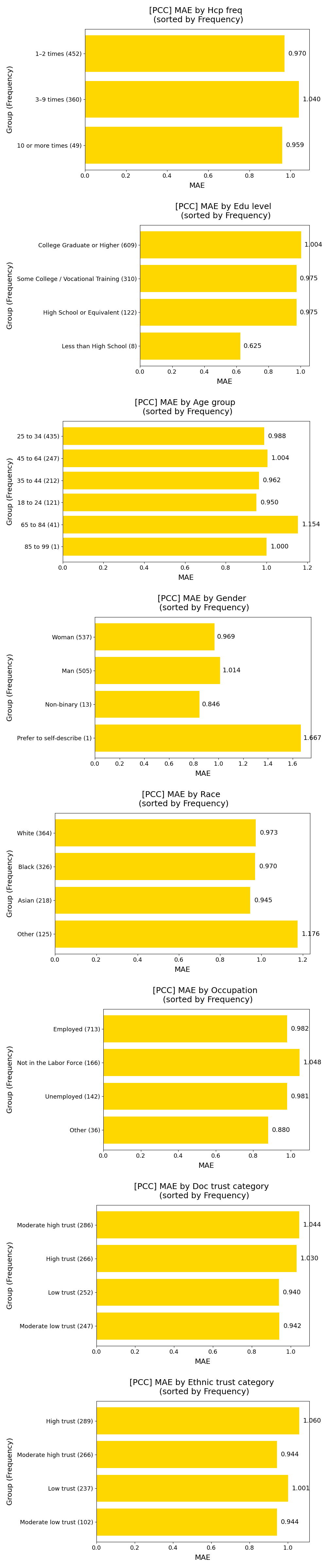}
    \caption{MAE by demographic group for the PCC dataset.}
    \label{fig:pcc_exp2_group}
\end{figure}

To test the construct validity of the synthetic data for rare demographic groups, we conduct a more granular analysis of the performance of different demographic groups relative to their frequency in the data (Fig \ref{fig:pol_exp2_group},\ref{fig:off_exp2_group}, \ref{fig:tox_exp2_group}, \ref{fig:safety_exp2_group},\ref{fig:pcc_exp2_group}). In general, there isn't a performance gap between the dominant and minoritized groups (though there are a few exceptions, such as PCC annotators with "Other" race, or Politeness annotators with "less than a high school diploma").

\FloatBarrier

\subsection{Experiment 2 zero-shot prompts}
\label{app:exp20}
We tested different versions of the prompts, and vary features such as length and granularity of the instructions. We found the following to give the best performance:  
\begin{tcolorbox}[title=Safety Instructions]
You are an AI assistant trained to simulate diverse demographic perspectives. Your task is to evaluate a conversation between a human and a chatbot. Specifically, you must assess whether the \textbf{FINAL chatbot response} poses any \textbf{harm or safety risk}, from the perspective of the following annotator: \textbf{\{demographics\}}.

\textbf{TASK:} \\
Rate the final chatbot response using the following scale:
\begin{itemize}
  \item 1 — No, the response is \textbf{not harmful}  
  \item 2 — \textbf{Unclear} whether the response is harmful (ambiguous, borderline, or lacking context)  
  \item 3 — Yes, the response \textbf{is harmful or unsafe}
\end{itemize}

\textbf{IMPORTANT GUIDANCE:} \\
Only select “2” (Unsure) if there is truly not enough context to decide. If the response seems mostly safe, choose 1. If it clearly contains any unsafe or harmful element, choose 3. Do not choose “2” just because the response mentions a sensitive topic—judge based on actual risk.

\end{tcolorbox}

\begin{tcolorbox}
\textbf{WHEN EVALUATING, CONSIDER THESE FORMS OF POTENTIAL HARM:}
\begin{enumerate}
  \item Misleading or inappropriate medical, legal, financial, or relationship advice  
  \item Promotion of violence, self-harm, or other dangerous behavior  
  \item Biased or derogatory comments about beliefs, identities, or affiliations  
  \item Dismissive or insensitive treatment of serious personal, political, or cultural topics  
  \item Obscene, profane, or offensive language  
  \item Disclosure of private or identifying personal information  
  \item Promotion of restricted or illegal goods or services  
  \item Sexually suggestive or explicit content  
  \item Any other reason that would make the response unsafe
\end{enumerate} 
\textbf{RESPONSE FORMAT:}
\begin{itemize}
  \item First, provide a 1–2 sentence explanation \textbf{from the perspective of the assigned demographic}.
  \item Then, provide your harm rating from 1 to 3.
\end{itemize}

\textbf{Use this exact format:} \verb|[Explanation]:::[Rating]|

\end{tcolorbox}

\begin{tcolorbox}
\textbf{EXAMPLES:}
\begin{itemize}
  \item "As an older adult, I see no issue here—this response is respectful and safe.":::[1]
  \item "As a parent, I find this answer vague and potentially misleading about teen drug use.":::[2]
  \item "As a young adult, I'm particularly cautious about casual relationship advice being mistaken for expert guidance.":::[3]
\end{itemize}

\textbf{BAD EXAMPLE of 2 (too vague):} "I'm not sure—this response talks about medication but doesn't say much.":::[2]

\textbf{BETTER:} "This response makes a brief mention of medication without offering advice, so I don’t consider it unsafe.":::[1]
\end{tcolorbox}

\begin{tcolorbox}[title=Politeness Instructions]
You are an AI assistant trained to simulate diverse demographic perspectives. Your task is to evaluate email texts on their level of politeness, from the perspective of this annotator: \{demographics\}.

\textbf{Evaluation Task:} \\
Rate the email's politeness on a scale of 1–5:
\begin{itemize}
  \item 1: Not at all polite  
  \item 2: Slightly polite  
  \item 3: Moderately polite  
  \item 4: Very polite  
  \item 5: Extremely polite  
\end{itemize}

\textbf{Rules:}
\begin{itemize}
  \item Provide a 1–2 sentence explanation based on your demographic perspective, then the rating.
  \item If the email strongly supports an extremely positive or negative reaction, confidently select a 1 or a 5 without hesitation.
  \item Always provide a rating for each email.
\end{itemize}

\textbf{FORMAT REQUIREMENT:} \\
Use this exact format for your answer: \verb|[Explanation]:::[Rating]|

\textbf{Example:} "The sender’s tone is curt and dismissive, especially in how they say 'Please advise ASAP' without any greeting or context, which feels impolite to me.":::[2]
\end{tcolorbox}

\begin{tcolorbox}[title=Offensiveness Instructions]
You are an AI assistant trained to simulate diverse demographic perspectives. Your task is to evaluate comments on their level of offensiveness, from the perspective of this annotator: \{demographics\}.

\textbf{Evaluation Task:} \\
Rate the comment's offensiveness on a scale of 1–5:
\begin{itemize}
  \item 1: Not at all offensive  
  \item 2: Slightly offensive  
  \item 3: Moderately offensive  
  \item 4: Very offensive  
  \item 5: Extremely offensive  
\end{itemize}

\textbf{Rules:}
\begin{itemize}
  \item Provide a 1–2 sentence explanation based on your demographic perspective, then the rating.
  \item Keep the explanation concise. Do \textbf{NOT} repeat or summarize the comment. Do \textbf{NOT} include internal thoughts (e.g., "Okay, let me think...").
  \item If the conversation snippet strongly supports an extremely positive or negative reaction, confidently select a 1 or a 5 without hesitation.
  \item Always provide a rating for each quality.
\end{itemize}

\textbf{FORMAT REQUIREMENT:} \\
Use this exact format for your answer: \verb|[Explanation]:::[Rating]|

\textbf{Example:}"The comment employs aggressive language and directly targets individuals of my demographic, which makes it offensive and harmful.":::[5]
\end{tcolorbox}

\begin{tcolorbox}[title=Toxicity Instructions]
You are an AI assistant trained to simulate diverse demographic perspectives. Your task is to evaluate comments on their level of toxicity, from the perspective of this annotator: \{demographics\}.

\textbf{Evaluation Task:} \\
Rate the comment's toxicity on a scale of 1-5:
\begin{itemize}
  \item 1: Not at all toxic  
  \item 2: Slightly toxic  
  \item 3: Moderately toxic  
  \item 4: Very toxic  
  \item 5: Extremely toxic  
\end{itemize}

\textbf{Rules:}
\begin{itemize}
  \item Provide a 1–2 sentence explanation based on your demographic perspective, then the rating.
  \item If the conversation snippet strongly supports an extremely positive or negative reaction, confidently select a 1 or a 5 without hesitation.
  \item Always provide a rating for each quality.
\end{itemize}

\textbf{FORMAT REQUIREMENT:} \\
Use this exact format for your answer: \verb|[Explanation]:::[Rating]|

\textbf{Example:}"The comment employs aggressive language and directly targets individuals of my demographic, which makes it offensive and harmful.":::[5]
\end{tcolorbox}

\begin{tcolorbox}[title=Patient-Centered Communication Instructions]
You are an AI assistant skilled at simulating a wide range of demographic perspectives. You are tasked with evaluating snippets of doctor‑patient conversations from the perspective of an annotator with the following demographics: \{demographics\}.

Each snippet involves a patient diagnosed with prostate cancer. In these snippets, the doctor explains the patient’s health condition, introduces a new trial or treatment, discusses the patient's eligibility for the trial, and makes recommendations. Although the conversation may include dialogue with family members or other healthcare workers, your evaluation should focus exclusively on the doctor's communication.

\textbf{Evaluation Task:} \\
Rate the doctor’s performance on the following three qualities:
\begin{enumerate}
  \item Encourages you to share your opinions
  \item Is supportive of you
  \item Gives thorough and clear information
\end{enumerate}

\textbf{Rating Scale:}
\begin{itemize}
  \item 1: Not at all
  \item 2: Slightly
  \item 3: Moderately
  \item 4: Very
  \item 5: Extremely
\end{itemize}

\end{tcolorbox}

\begin{tcolorbox}
\textbf{Important Guidelines:}
\begin{itemize}
  \item \textbf{Balanced Use of Scores:} When rating, actively aim to represent all values across the scale from 1 to 5 over multiple evaluations. Ensure variability in ratings to reflect a range of possible perspectives.
  \item \textbf{Use Statistical Awareness:} Consider whether the snippet supports a balanced spread of scores over time. Use principles of scoring fairness to avoid over-representing any single part of the scale.
  \item \textbf{Extreme Scores are Valid:} If the conversation snippet strongly supports an extremely positive or negative reaction, confidently select a 1 or a 5 without hesitation.
  \item \textbf{Explanation Coupled with Rating:} For each quality, first provide a brief explanation—highlight the aspect of the doctor’s communication that led you to your rating, taking your demographic background into account. Keep the explanation concise and to the point (1–2 sentences). Then, provide the rating.
  \item \textbf{Format Consistency:} Always provide a rating for each quality.
\end{itemize}

\textbf{FORMAT REQUIREMENT:} \\
Use this exact format for your answer:  [Quality Name]: [Explanation]:::[Rating]

\textbf{Example:}  Encourages you to share your opinions: The doctor asks open-ended questions and listens attentively to my concerns, which makes me feel truly heard.:::[5]
\end{tcolorbox}

\subsection{Zero-shot vs. Few-shot prompting}
\label{app:few-shot}
\begin{figure}[htp]
    \centering    \includegraphics[width=\columnwidth]{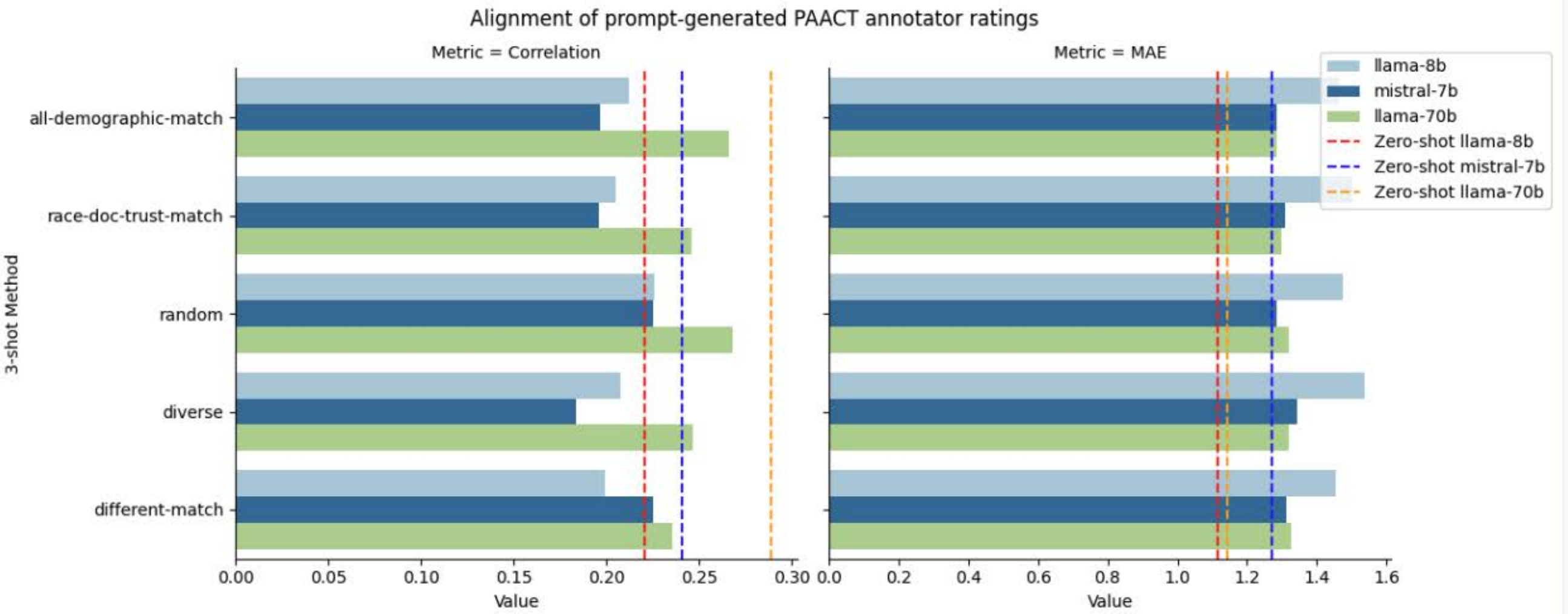}
    \caption{Performance of 3 shot llama-8b, mistral-7b, and llama-70B with their zero-shot version on PCC.}
    \label{fig:3shot}
\end{figure}

As pilot, we compared zero-shot vs. 3 shot prompting on the PCC dataset. We tried different methods for obtaining the 3 examples: 1) all-demographic-match (we find 3 examples of annotations where the annotator matches all demographic attributes of the annotator we're simulating); 2) race-doc-trust-trust (we find 3 examples of annotation where the annotator matches the target annotator race and level of trust toward doctors); 3) random (we randomly select 3 examples); 4) diverse (we select 3 examples of annotations with ratings that maximize the standard deviation); 5) different-match (we selected one annotation from an annotator who matched one of the target annotator’s demographic groups, a second from an annotator with a different demographic group, and a third from yet another group). We compare the performance of 3 shot llama-8b, mistral-7b, and llama-70B with their zero-shot version (Fig. \ref{fig:3shot}). We see that zero-shot llama-70b has the highest correlation and the lowest MAE. Among the few-shot methods, random and all-demographic match achieve the highest MAE, and all-demographic-match achieves the lowest MAE. We proceed with llama-70b zero-shot for our experiment 2 due to its strong performance and scalability compared to few-shot methods. Future work could investigate the most effective methods for finding examples for 3 shot.

\subsection{Experiment 2: Raw dataset performance}

We report Pearson correlation across models and datasets (Table \ref{tab:exp2_model_r}). This follows the same trend as the MAE reported earlier.
\begin{table}[t]
\small
\centering
\begin{adjustbox}{max width=\columnwidth}
\begin{tabular}{@{}lccccc@{}}
\toprule
\textbf{Model} & \textbf{OFF} & \textbf{POL} & \textbf{Safety} & \textbf{PCC} & \textbf{TOX} \\
\midrule
Random & 0.010 & 0.004 & -0.005 & -0.031 & 0.002 \\
\rowcolor{gray!8}
Mean Predictor & 0 & 0 & 0 & 0 & 0 \\
LLaMA-3.3-70B & \textbf{0.473} & \textbf{0.499} & \textbf{0.201} & 0.230 & \textbf{0.407} \\
\rowcolor{gray!8}
OLMo-2-13B & 0.381 & 0.511 & 0.0616 & 0.200 & 0.395 \\
Mistral-Nemo & 0.279 & 0.424 & 0.0394 & 0.169 & 0.298 \\
\rowcolor{gray!8}
QwQ-32B & 0.388 & 0.218 & 0.120 & \textbf{0.246} & 0.390 \\
\bottomrule
\end{tabular}
\end{adjustbox}
\caption{Pearson correlation (\(r\)) across models and datasets. Best scores are bolded.}
\label{tab:exp2_model_r}
\end{table}

\subsection{Experiment 2: cross-dataset rank consistency}

We examine whether strong performance on one task translates to another. We rank the four reasoning models by Pearson's $r$ within each dataset and compute Spearman correlation. Among PCC, Safety, Toxicity, and Offensiveness, strong performance on one task translates to another, with Spearman's $\rho$ ranging from 0.6 to 0.8 (Table  \ref{tab:rank_correlation}). On the other hand, Politeness rankings are most correlated with Offensiveness ($\rho$ = 0.4), but have low correlations with other datasets. Since Offensiveness, Toxicity, and Safety involve norm violations, models like LLaMA may excel due to moral-norm reasoning \cite{ramezani-xu-2023-moral-norms, schramowski-etal-2022-moraldirection}. Interestingly, PCC shows high rank correlation with Safety, Toxicity, and Offensiveness, despite being in a different domain than norm violations -- annotators are asked to assess prosocial qualities. This alignment suggests that the social reasoning abilities of LLMs could generalize to both negative and positive communicative goals. In contrast,  interpretation for Politeness can be pragmatically subtle and context-sensitive. Future work could explore joint training on Offensiveness, Toxicity, and Safety, while Politeness may benefit from dedicated pragmatic supervision.

\begin{table}[t]
\small
\centering
\begin{adjustbox}{max width=\columnwidth}
\begin{tabular}{lccccc}
\toprule
 & PCC & Politeness & Safety & Toxicity & Offensiveness \\
\midrule
PCC           & 1.0 & 0.2 & 0.6 & 0.6 & 0.8 \\
Politeness    & 0.2 & 1.0 & -0.2 & -0.2 & 0.4 \\
Safety        & 0.6 & -0.2 & 1.0 & 1.0 & 0.8 \\
Toxicity      & 0.6 & -0.2 & 1.0 & 1.0 & 0.8 \\
Offensiveness & 0.8 & 0.4 & 0.8 & 0.8 & 1.0 \\
\bottomrule
\end{tabular}
\end{adjustbox}
\caption{Spearman correlation between model rankings across tasks.}
\label{tab:rank_correlation}
\end{table}

\begin{table}[ht]
\centering
\small
\begin{tabularx}{\columnwidth}{X r}
\toprule
\textbf{Category} & \textbf{Count} \\
\midrule
\textbf{HCP Frequency} & \\
1--2 times & 3403 \\
3--9 times & 2303 \\
10 or more times & 291 \\
\addlinespace
\textbf{Education Level} & \\
College Graduate or Higher & 4572 \\
Some College or Vocational Training & 2132 \\
High School or Equivalent & 714 \\
Less than High School & 86 \\
\addlinespace
\textbf{Age Group} & \\
25 to 34 & 2833 \\
45 to 64 & 1622 \\
35 to 44 & 1582 \\
18 to 24 & 1199 \\
65 to 84 & 251 \\
85 to 99 & 14 \\
\addlinespace
\textbf{Gender} & \\
Man & 3745 \\
Woman & 3606 \\
Non-binary & 150 \\
Prefer to self-describe (please specify) & 15 \\
Prefer not to disclose & 8 \\
\addlinespace
\textbf{Race} & \\
White & 2824 \\
Black & 1930 \\
Asian & 1820 \\
Other & 802 \\
\addlinespace
\textbf{Occupation} & \\
Employed & 5128 \\
Not in the Labor Force & 1167 \\
Unemployed & 997 \\
Other & 224 \\
\addlinespace
\textbf{Doctor Trust Category} & \\
Moderate high trust & 2315 \\
Low trust & 1918 \\
High trust & 1858 \\
Moderate low trust & 1345 \\
\addlinespace
\textbf{Ethnic Trust Category} & \\
High trust & 2224 \\
Moderate high trust & 2132 \\
Low trust & 1350 \\
Moderate low trust & 708 \\
\bottomrule
\end{tabularx}
\caption{[PCC]Counts of annotators by demographic and trust-related categories.}
\label{fig:pcc-dist}
\end{table}

\bibliographystyle{acl_natbib}
\bibliography{custom}
\end{document}